\pdfoutput=1

\documentclass[11pt]{article}

\usepackage[final]{acl}

\usepackage{times}
\usepackage{latexsym}

\usepackage[T1]{fontenc}

\usepackage[utf8]{inputenc}

\usepackage{microtype}

\usepackage{inconsolata}

\usepackage{graphicx}

\usepackage{xspace}
\usepackage{amsmath}
\usepackage{mathtools}
\usepackage{booktabs, nicematrix}
\usepackage{multirow}
\usepackage{tikz}
\usepackage{fontawesome}
\usepackage{CJKutf8}
\usepackage{xcolor}
\usepackage{color, colortbl}
\usepackage{siunitx}
\usepackage{fixfoot}
\usepackage{hyperref}
\usepackage{tablefootnote}

\definecolor{highlight}{gray}{0.90}

\newcommand{\good}{\colorbox{green!10}{\textsc{Good}}\xspace}
\newcommand{\bad}{\colorbox{red!10}{\textsc{Bad}}\xspace}
\newcommand{\perfect}{\colorbox{blue!10}{\textsc{Perfect}}\xspace}
\newcommand{\other}{\colorbox{highlight}{\textsc{Other}}\xspace}

\newcommand{\gemba}{\textsc{gemba-mqm}\xspace}
\newcommand{\matese}{MaTESe\xspace}
\newcommand{\mateseqe}{MaTESe-QE\space}
\newcommand{\comet}{\textsc{comet}\xspace}
\newcommand{\cometqe}{\textsc{comet-qe}\xspace}
\newcommand{\cometmqmqe}{\textsc{comet-qe-mqm}\xspace}
\newcommand{\xcomet}{x\textsc{comet}\xspace}
\newcommand{\xcometens}{x\textsc{comet-ensemble}\xspace}
\newcommand{\xcometensqe}{x\textsc{comet-qe-ensemble}\xspace}
\newcommand{\xcometxl}{x\textsc{comet-xl}\xspace}

\newcommand{\metricx}{MetricX-23\xspace}
\newcommand{\metricxqe}{MetricX-23-QE\xspace}
\newcommand{\metricxxl}{MetricX-23-XL\xspace}
\newcommand{\metricxxlqe}{MetricX-23-QE-XL\xspace}
\newcommand{\mbrmetricxqe}{MBR-MetricX-QE\xspace}
\newcommand{\bleurt}{\textsc{bleurt-20}\xspace}
\newcommand{\cometkiwi}{CometKiwi\xspace}
\newcommand{\cometkiwixl}{CometKiwi-XL\xspace}
\newcommand{\dasqm}{DA+SQM\xspace}
\newcommand{\randombaseline}{Random-sysname\xspace}

\newcommand{\bertscore}{BERTscore\xspace}
\newcommand{\bleu}{BLEU\xspace}
\newcommand{\ebleu}{eBLEU\xspace}
\newcommand{\fbleu}{f200spBLEU\xspace}
\newcommand{\chrf}{chrF\xspace}
\newcommand{\tokengramf}{tokengram\_F\xspace}

\newcommand{\candonly}{$\textsc{sentinel}_{\textsc{cand}}$\xspace}
\newcommand{\srconly}{$\textsc{sentinel}_{\textsc{src}}$\xspace}
\newcommand{\refonly}{$\textsc{sentinel}_{\textsc{ref}}$\xspace}

\newcommand{\langpair}[2]{\textsc{#1}$\rightarrow$\textsc{#2}}
\newcommand{\bs}[1]{\boldsymbol{#1}}
\newcommand{\acceq}{$\text{acc}_{\text{eq}}$\xspace}

\newcommand{\pme}{P^{\mathcal{M}_{\tau}}}
\newcommand{\rme}{R^{\mathcal{M}_{\tau}}}
\newcommand{\fme}{F^{\mathcal{M}_{\tau}}}
\newcommand{\fscore}{$F$-score\xspace}
\newcommand{\wmttw}{WMT22$_{\text{MQM}}$\xspace}
\newcommand{\wmtth}{WMT23$_{\text{MQM}}$\xspace}
\newcommand{\wmtthda}{WMT23$_{\text{DA+SQM}}$\xspace}

\title{Beyond Correlation: Interpretable Evaluation of Machine Translation Metrics}

\author{Stefano Perrella\textsuperscript{*} \quad Lorenzo Proietti\textsuperscript{*} \quad  Pere-Lluís Huguet Cabot \\ {\bf Edoardo Barba} \quad {\bf Roberto Navigli} \\ Sapienza NLP Group, Sapienza University of Rome \\ \small \texttt{\{perrella, lproietti, huguetcabot, barba, navigli\}@diag.uniroma1.it}}

\begin{document}
\maketitle

\def\thefootnote{*}\footnotetext{Equal contribution.}
\def\thefootnote{\arabic{footnote}}

\begin{abstract}
Machine Translation (MT) evaluation metrics assess translation quality automatically. Recently, researchers have employed MT metrics for various new use cases, such as data filtering and translation re-ranking. However, most MT metrics return assessments as scalar scores that are difficult to interpret, posing a challenge to making informed design choices. Moreover, MT metrics' capabilities have historically been evaluated using correlation with human judgment, which, despite its efficacy, falls short of providing intuitive insights into metric performance, especially in terms of new metric use cases. To address these issues, we introduce an interpretable evaluation framework for MT metrics. Within this framework, we evaluate metrics in two scenarios that serve as proxies for the data filtering and translation re-ranking use cases. Furthermore, by measuring the performance of MT metrics using Precision, Recall, and $F$-score, we offer clearer insights into their capabilities than correlation with human judgments. Finally, we raise concerns regarding the reliability of manually curated data following the Direct Assessments+Scalar Quality Metrics (DA+SQM) guidelines, reporting a notably low agreement with Multidimensional Quality Metrics (MQM) annotations.

\end{abstract}

\begin{figure}[t]
    \centering
    \resizebox{\columnwidth}{!}{
    \includegraphics{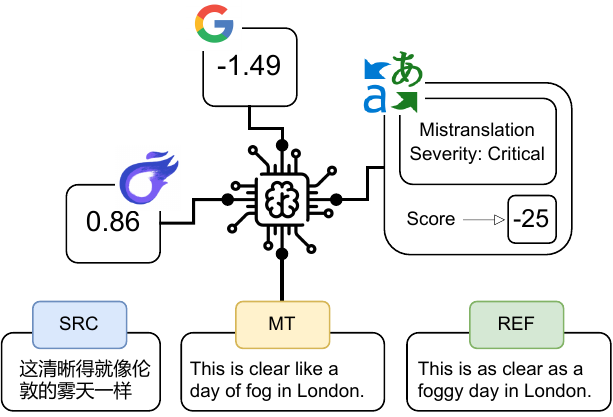}
    }
    \caption[]{Quality assessments returned by \comet \cite{rei-etal-2020-comet}, \metricxxlqe \cite{juraska-etal-2023-metricx}, and \gemba \cite{kocmi-federmann-2023-gemba} for the provided machine-translated text.}
    \label{fig:figure1}
\end{figure}

\section{Introduction}
Over the past few years, Machine Translation (MT) evaluation metrics have transitioned from heuristic-based to neural-based, enabling a more nuanced evaluation of translation quality and a greater agreement with human judgments \cite{freitag-etal-2022-results}. Additionally, recent Metrics Shared Tasks at the Conference on Machine Translation \cite[WMT]{mathur-etal-2020-results, freitag-etal-2021-results} have seen the rise of reference-free metrics, which assess translation quality without the need for human-curated references by comparing translations only to their sources in the original language. Lately, reference-free metrics have demonstrated performance on par with, and sometimes superior to, their reference-based counterparts \cite{freitag-etal-2023-results, kocmi-etal-2024-navigating}. Thanks to these advancements and the ability to use metrics without references, several new MT metrics use cases have emerged. 
\citet{freitag-etal-2022-high}, \citet{fernandes-etal-2022-quality}, \citet{farinhas-etal-2023-empirical}, \citet{ramos-etal-2024-aligning}, and \citet{finkelstein2024mbr} used MT metrics as utility functions for Minimum Bayes Risk (MBR) decoding \cite{kumar-byrne-2004-minimum, eikema-aziz-2020-map} and for Quality Estimation (QE) re-ranking.\footnote{MBR decoding and QE re-ranking are methods used to identify the best translation from multiple outputs generated by an MT system for the same source text. MBR decoding typically relies on reference-based metrics, while QE re-ranking depends on reference-free metrics.}
\citet{ramos-etal-2024-aligning}, \citet{gulcehre2023reinforced}, \citet{he-etal-2024-improving}, and \citet{xu2024contrastive} used MT metrics as a proxy for human preferences to fine-tune MT systems using Reinforcement Learning (RL)- and Direct Preference Optimization (DPO)-like training objectives.
\citet{peter-etal-2023-theres}, \citet{alves2024tower}, and \citet{gulcehre2023reinforced} used reference-free metrics to filter parallel corpora -- discarding all translations assigned with a metric score that is below a certain threshold -- with the goal of training MT systems using higher quality data.
These works leverage MT metrics for applications beyond their traditional use of measuring incremental improvements in the development of MT systems. However, the lack of a dedicated evaluation, paired with the inherent opacity of MT metrics, makes it challenging to determine whether one metric suits a given task better and what the impact of various design choices is.
For example, \citet{alves2024tower}, \citet{peter-etal-2023-theres}, and \citet{gulcehre2023reinforced} filter MT datasets using different MT metrics and thresholds, leaving it unclear whether an optimal choice exists. Furthermore, considering the ever-increasing number of metrics available, researchers are often limited to grid-searching for the best configuration for each new application, as do \citet{fernandes-etal-2022-quality} and \citet{ramos-etal-2024-aligning}, who explore by grid-search whether certain metrics are better suited than others for MBR decoding, QE re-ranking, and as reward models for RL-based training. 
However, the lack of dedicated evaluation setups often requires revisiting these studies to assess whether their findings hold with newer metrics, resulting in a non-negligible increase in experimentation time.

In this work, we address these issues by introducing a novel and more interpretable evaluation framework for MT metrics, comprising evaluation setups designed as proxies for new metric use cases. In the following sections, we first illustrate the problem of interpretability, then introduce our framework, and finally present our results.

\section{The Interpretability of MT Metrics' Assessments} \label{seq:interpretability}

In the field of AI, Interpretability is defined as 
``the ability to explain or to provide the meaning in understandable terms to a human'' \cite{arrieta2019explainable}, and typically refers to the problem of understanding the decision-making process of an AI model. However, our goal here is less ambitious. 
Instead of focusing on the interpretability of MT metrics themselves, we are concerned with the interpretability of their assessments. Specifically, most state-of-the-art MT metrics are trained to minimize the Mean Squared Error (MSE) with human judgments and return assessments as scalar quality scores, which are difficult to interpret. Therefore, we are interested in understanding the meaning of these scores, rather than the internal workings of MT metrics. 

In light of this, we attribute MT metrics assessments' lack of interpretability to three main factors:\footnote{We wish to clarify that we identified these three factors as notably impactful, but they are non-exhaustive and may overlap.}
\begin{enumerate}
    \item \textbf{Range consistency}: it is unclear whether a difference in metric score has the same meaning if it occurs in different regions of the score range.

    \item \textbf{Error attribution}: scalar quality assessments do not identify specific translation errors.
    
    \item \textbf{Performance}: metrics capabilities are typically measured through correlation with human judgment, which fails to provide users with a clear understanding of their performance and reliability.
\end{enumerate}
In simpler terms, let us consider the example of Figure~\ref{fig:figure1}.
Due to the lack of \textit{Error attribution} we do not know which translation errors, if any, led \comet \cite{rei-etal-2020-comet} to return $0.86$. Also, the metric comes with no \textit{Range consistency} guarantee, e.g. whether $0.86$ is twice as good as $0.43$. Furthermore, different metrics have different score ranges, making it difficult to compare the assessments from \comet with those of the other MT metrics in the figure. Finally, lacking a clear understanding of \comet's \textit{Performance} beyond human correlation, we cannot be sure whether we can draw conclusions from its assessments safely.

For this reason, some efforts have been made to design interpretable metrics. For example, among the primary submissions to the WMT23 Metrics Shared Task \cite{freitag-etal-2023-results}, \matese \cite{perrella-etal-2022-matese} annotates the spans of a translation that contain errors, specifying their severity, \xcomet models \cite{guerreiro2023xcomet} return annotated error spans together with a final regression value, and \gemba \cite{kocmi-federmann-2023-gemba} leverages GPT-4 \cite{openai2024gpt4} to produce detailed quality assessments. However, these metrics compromise on other aspects to accommodate the increased interpretability. \matese displays a lower correlation with human judgment than several regression-based metrics.\footnote{While this might be due to several contributing factors, the limited availability of training data containing detailed span-level annotations is most likely one of them.} Trading off performance and interpretability, \xcomet models' final assessment is based mainly on the regression component, with annotated spans contributing only $2/9$ of the overall score. Finally, \gemba is prohibitively expensive to operate and its assessments are not fully-reproducible, due to its dependence on the closed-source GPT-4. 

In this work, we seek to mitigate the interpretability issue by targeting the problem of \textit{Performance}. Specifically, we take inspiration from two popular new MT metrics applications, i.e., data filtering and translation re-ranking, in order to study and measure metrics performance in terms of Precision, Recall, and $F$-score. Taking advantage of these measures that are more transparent than correlation, we aim to shed light on the meaning and reliability of metrics assessments, especially concerning such new MT metrics use cases. We release our evaluation framework as software at \url{https://github.com/SapienzaNLP/interpretable-mt-metrics-eval}.

\section{An Interpretable Evaluation Framework for MT Metrics}
Two popular new MT metrics applications are data filtering and translation re-ranking. In data filtering, MT metrics separate good-quality from poor-quality translations. After choosing a threshold value, all translations below the threshold are labeled as poor quality and discarded. 
In this respect, we are interested in jointly assessing metric performance and studying the meaning of metric scores, finding the thresholds that best separate good-quality (\good) from poor-quality (\bad) translations.
Instead, in translation re-ranking, MT metrics determine the best in a pool of translations of the same source text. For example, in QE re-ranking and MBR decoding, metrics are tasked to identify the best translation among those sampled from an MT system. 

With the aim of facilitating practitioners in making design choices for these metrics applications, and with a focus on the interpretability issue, we evaluate MT metrics performance in two settings: i) when metrics are used as binary classifiers, tasked to separate between \good and \bad translations (acting as a proxy for the data filtering application), and ii) when metrics are used to identify the best translation in a group of translations of the same source (acting as a proxy for translation re-ranking).

\subsection{Metrics as Binary Classifiers for Data Filtering} \label{seq:binary-classifiers}
Let us consider the MT metric $\mathcal{M}$, which outputs scores in the range $[m_1,m_2]$. Let us define $\mathcal{M}(t) \in [m_1, m_2]$ as the score assigned by metric $\mathcal{M}$ to translation $t$.
By selecting an arbitrary threshold value $\tau \in [m_1, m_2]$, we repurpose $\mathcal{M}$ as a binary classifier: a translation $t$ is deemed as \good by metric $\mathcal{M}$, with threshold $\tau$, if $\mathcal{M}(t) \geq \tau$, otherwise it is deemed as \bad.

\paragraph{Precision, Recall, and $\boldsymbol{F}$-score}
Assuming that we have an oracle $\mathcal{H}$ telling us whether a translation is \good or \bad, we can measure the performance of metric $\mathcal{M}$, with threshold $\tau$, in terms of standard measures such as Precision, Recall, and \fscore, which we refer to as $\pme$, $\rme$, and $\fme$.
Given metric $\mathcal{M}$, oracle $\mathcal{H}$, translation $t$, and threshold $\tau$, $\pme$ estimates the probability that translation $t$ is \good, given that metric $\mathcal{M}$ deems it as such:
\begin{equation}
    \pme = \hat{\Pr}(\mathcal{H}(t) = \text{\good} \mid \mathcal{M}(t) \geq \tau).
    \label{eq:prob_precision}
\end{equation}
Similarly, $\rme$ estimates the probability that metric $\mathcal{M}$ deems translation $t$ as \good, given that the oracle deems it as such:
\begin{equation}
    \rme = \hat{\Pr}(\mathcal{M}(t) \geq \tau \mid \mathcal{H}(t) = \text{\good}).
    \label{eq:prob_recall}
\end{equation}
Finally, we aggregate Precision and Recall using $F_{\beta}$-score, with $\beta = \frac{1}{\sqrt{2}}$, which weights Precision higher than Recall compared to the more common $F_1$-score. Arguably, false positives -- i.e., translations of low quality that are mistakenly considered \good -- could be detrimental to the applications that see metrics employed as binary classifiers. For example, in data filtering, false positives correspond to low-quality translations that survive the filtering, compromising the quality of filtered data. In contrast, false negatives -- i.e., translations of high quality that are mistakenly assigned with the \bad label -- would more frequently lead to minor inconveniences, as they correspond to good-quality translations that are mistakenly discarded.  Moreover, we note that MT metrics struggle to achieve high Precision, meaning that metrics differences can be best highlighted if Precision is weighted higher than Recall.

Therefore, the $F$-score of metric $\mathcal{M}$, with threshold $\tau$, is defined as follows: 
\begin{equation}
    \fme = \frac{3}{2} \frac{\pme \rme}{\frac{1}{2} \pme + \rme}.
    \label{eq:fscore}
\end{equation}


\subsection{Metrics as Utility Functions for Translation Re-Ranking}
Let us consider the set $T = \{t_1, t_2, ..., t_n\}$ containing translations of the same source text. 
We are interested in assessing metric performance in ranking the best translation, as determined by human annotators, in the first position. However, metrics and humans might return tied assessments, placing two or more translations together in the first position. Therefore, we define $T^{\mathcal{M}}$ as the subset of $T$ containing all translations assigned with the highest score by $\mathcal{M}$. Similarly, $T^{\mathcal{H}}$ contains the translations of $T$ ranked highest by human annotators.
The Re-Ranking Precision of metric $\mathcal{M}$ is defined as follows:
\begin{equation}
    RRP^{\mathcal{M}} = \frac{|T^{\mathcal{M}} \cap T^{\mathcal{H}}|}{|T^{\mathcal{M}}|}. \label{eq:accuracy}
\end{equation}

Unlike in the data filtering scenario, we focus solely on Re-Ranking Precision, not Recall. This is because, to serve as a proxy for translation re-ranking applications, what matters is whether the returned translation is the best -- or among the best -- rather than identifying all the translations ranked highest by human annotators.

\section{Experimental Setup}
This section outlines the data employed, the metrics evaluated, and our methodology. Implementation details regarding the calculation of Precision, Recall, and \fscore are in Appendix~\ref{apx:implementation-details}.

\subsection{The Data}
We employ \textbf{\wmtth} \cite{freitag-etal-2023-results}, which contains human annotations collected within the Multidimensional Quality Metrics framework \cite[MQM]{mqm-framework}, and \textbf{\wmtthda} \cite{kocmi-etal-2023-findings}, which includes human annotations as Direct Assessments + Scalar Quality Metrics \cite[DA+SQM]{kocmi-etal-2022-findings}. 
Both datasets consist of source texts translated by multiple MT systems, with translation quality assessed by professional human annotators. Table~\ref{tab:datastats} in the Appendix provides additional information regarding these datasets.

We conduct the evaluation using \wmtth, specifically the \langpair{zh}{en} language direction, and use \wmtthda as a reference for human performance, given that it contains a subset of the translations in \wmtth annotated using a different human evaluation scheme. Results concerning the other language directions in \wmtth, i.e., \langpair{en}{de} and \langpair{he}{en}, are reported in the Appendix.

\subsection{The Metrics}
We consider the following metrics: \comet, \cometqe, and \cometmqmqe \cite{rei-etal-2020-comet, rei-etal-2021-references};
\bleurt \cite{sellam-etal-2020-bleurt, pu-etal-2021-learning}; 
\metricx, \metricxqe, \metricxxl, and \metricxxlqe \cite{juraska-etal-2023-metricx}; 
\cometkiwi and \cometkiwixl \cite{rei-etal-2022-cometkiwi, rei-etal-2023-scaling}; 
\xcometens and \xcometensqe \cite{guerreiro2023xcomet};
\xcometxl \cite{guerreiro2023xcomet};  
\matese and \mateseqe \cite{perrella-etal-2022-matese};  
\gemba \cite{kocmi-federmann-2023-gemba}; 
\mbrmetricxqe \cite{naskar-etal-2023-quality}.
We refer the reader to Appendix~\ref{apx:metrics} for detailed information regarding these metrics, where we also provide a broader selection, including lexical-based and sentinel metrics \cite{perrella-etal-2024-guardians}. 

Additionally, following \citet{freitag-etal-2023-results}, we include the results from a random baseline, i.e., \randombaseline, which outputs discrete scores drawn from several Gaussian distributions, one for each MT system that translated the texts in the test set. Each Gaussian has a randomly assigned mean between $0$ and $9$, with a standard deviation of $2$. 

\subsection{Selecting the Thresholds \texorpdfstring{$\bs{\tau}$}{tau}}
In the data filtering scenario, we can measure two different aspects of metric performance, depending on how we select the $\tau$ value:
\begin{enumerate}
    \item By selecting $\tau$ to maximize the \fscore \textbf{on the test set}, we measure MT metrics' ability to separate \good from \bad translations under ideal conditions. This scenario allows us to measure the maximum achievable \fscore for each metric on the test data, effectively evaluating the metric's discriminative power. Metrics whose assessments are not accurate enough, noisy, or, more generally, poorly aligned with human judgments, will achieve a lower optimal \fscore than the others.

    \item By selecting $\tau$ to maximize the \fscore \textbf{on a development set}, we estimate the true values of MT metrics' Precision, Recall, and \fscore in data filtering applications.
\end{enumerate}
We measure metric performance in both evaluation scenarios. As a development set, we use \textbf{\wmttw}, which contains MQM-based human annotations \cite{freitag-etal-2022-results}. However, since some of the tested metrics were trained using \wmttw data, we restrict this experiment to the other metrics. 

\subsection{Extracting Binary Labels from Manually-Annotated Datasets}
Within the MQM annotation framework, professional annotators identify span-level translation errors and assign each error a category and severity. The final MQM score is calculated based on these errors using the following weighting scheme \cite{freitag-etal-2021-experts}:

\begin{center}
    \resizebox{0.8\columnwidth}{!}{
    \begin{tabular}{llr}
        Error severity & Category & \multicolumn{1}{l}{Penalty} \\
        \cmidrule(lr){1-3}
        \multirow{2}{*}{Major} & Non-translation & $-25$ \\
        & Others & $-5$ \\
        \multirow{2}{*}{Minor} & Punctuation &  $-0.1$ \\
        & Others &  $-1$ \\
    \end{tabular}
    }
\end{center}

We map the annotations in \wmtth and \wmttw to binary labels by considering translations with a score above a certain threshold as \good. Specifically, if a translation is assigned an MQM score $h$, we label it as \good if $h \ge -4$, meaning it contains no Major errors and at most $4$ Minor ones (or more if Minor punctuation errors are present). Additionally, we classify translations as \perfect if they contain at most $1$ Minor error, i.e., those with $h \ge -1$.\footnote{We use $h \ge -1$ and not $h = 0$ because the inter-annotator agreement in MT evaluation is not particularly high \cite{freitag-etal-2021-experts}, even with high-cost annotation frameworks like MQM. Therefore, we argue that selecting only translations with a score of $0$ might overly depend on individual annotators' preferences.} This allows us to investigate metrics' ability to distinguish between \perfect and \other translations.

\section{Results}
In this section, we report the performance obtained by MT metrics when used as binary classifiers to distinguish between \good and \bad, as well as \perfect and \other translations, and in terms of their effectiveness in translation re-ranking, i.e., in selecting the best translations among candidates for the same source text.

\subsection{Binary Classification} \label{sec:results-data-filtering}
Table~\ref{tab:performance-results-zhen} shows MT metrics' threshold values, Precision, Recall, and \fscore in distinguishing \good from \bad and \perfect from \other translations, with the optimal threshold $\tau$ selected on the test set. 
As can be seen, most MT metrics perform reasonably well in distinguishing between \good and \bad translations, achieving optimal $F$-scores as high as $81.59$ and $81.40$, from \gemba and \xcometensqe, respectively, and as low as $75.81$, from \bleurt. Instead, lower performance is observed when differentiating between \perfect and \other translations, with the highest \fscore being $68.47$, from \xcometens. 
We also note that Precision is almost always lower than Recall, despite the optimal threshold $\tau$ being selected to maximize $F_{\beta}$-score with $\beta=\frac{1}{\sqrt{2}}$, which gives more weight to Precision over Recall. These results suggest that the metrics may lack the sensitivity required to distinguish between high-quality translations that differ in minor nuances rather than major errors. As a result, they may resort to lower thresholds, compensating for their lack of Precision with a higher Recall. 

Table~\ref{tab:performance-results-zhen-dev} reports threshold values, Precision, Recall, and \fscore when the threshold is optimized on the development set. Note that we restrict the set of tested metrics to those that are openly available and do not employ the \wmttw data for training. As expected, the \fscore values are lower than the optimal ones reported in Table~\ref{tab:performance-results-zhen}. Nonetheless, the metric rankings remain stable across the two settings, with \metricxxl and \metricxxlqe outperforming the other metrics among the reference-based and reference-free ones, respectively.

In general, it is worth noting that the best-performing openly available, reference-free metric is \metricxxlqe. This result is consistent across language pairs (Appendix~\ref{apx:performance-results}). 
Therefore, we recommend using \metricxxlqe for data filtering applications.

\begin{table*}[t]
\centering
\resizebox{\linewidth}{!}{
    \begin{NiceTabular}{ll|rrrr|rrrr|rr}[cell-space-limits=3pt]
    \toprule
     &  & \multicolumn{4}{c|}{\textbf{\good vs \bad}} & \multicolumn{4}{c|}{\textbf{\perfect vs \other}} & \multicolumn{2}{c}{\textbf{Re-ranking}} \\
    & \textbf{Metric} & \multicolumn{1}{c}{$\boldsymbol{\tau}$} & \multicolumn{1}{c}{\textbf{P}} & \multicolumn{1}{c}{\textbf{R}} & \multicolumn{1}{c|}{$\boldsymbol{F}$} & \multicolumn{1}{c}{$\boldsymbol{\tau}$} & \multicolumn{1}{c}{\textbf{P}} & \multicolumn{1}{c}{\textbf{R}} & \multicolumn{1}{c|}{$\boldsymbol{F}$} & \multicolumn{1}{c}{\textbf{RRP}} & \multicolumn{1}{c}{\textbf{Avg.}} \\
    \cmidrule(lr){2-12}
    \multirow{ 7 }{*}{\rotatebox{90}{\shortstack{\textsc{reference} \\ \textsc{based}}}}
    & \cellcolor{highlight}\xcometens & \cellcolor{highlight}$0.83$ & \cellcolor{highlight}$79.91$ & \cellcolor{highlight}$84.42$ & \cellcolor{highlight}$81.36$ & \cellcolor{highlight}$0.91$ & \cellcolor{highlight}$68.25$ & \cellcolor{highlight}$68.93$ & \cellcolor{highlight}$68.47$ & \cellcolor{highlight}$43.17$ & \cellcolor{highlight}$-2.38$ \\
    & \xcometxl & $0.80$ & $78.33$ & $83.63$ & $80.02$ & $0.92$ & $67.55$ & $67.46$ & $67.52$ & $37.49$ & $-2.75$ \\
    & \cellcolor{highlight}\metricx & \cellcolor{highlight}$-4.79$ & \cellcolor{highlight}$77.43$ & \cellcolor{highlight}$86.23$ & \cellcolor{highlight}$80.15$ & \cellcolor{highlight}$-2.25$ & \cellcolor{highlight}$63.99$ & \cellcolor{highlight}$73.20$ & \cellcolor{highlight}$66.79$ & \cellcolor{highlight}$39.63$ & \cellcolor{highlight}$-2.72$ \\
    & \metricxxl & $-3.52$ & $77.80$ & $84.46$ & $79.90$ & $-1.74$ & $65.60$ & $72.54$ & $67.76$ & $39.52$ & $-2.71$ \\
    & \matese & $-4.00$ & $76.53$ & $78.10$ & $77.05$ & $-1.00$ & $55.75$ & $79.88$ & $61.99$ & $33.07$ & $-3.18$ \\
    & \comet & $0.76$ & $74.56$ & $78.76$ & $75.91$ & $0.82$ & $61.25$ & $64.38$ & $62.26$ & $34.25$ & $-3.06$ \\
    & \bleurt & $0.60$ & $72.76$ & $82.76$ & $75.81$ & $0.67$ & $55.88$ & $69.21$ & $59.71$ & $33.35$ & $-3.07$ \\
    \cmidrule(lr){2-12}
    \multirow{ 10 }{*}{\rotatebox{90}{\shortstack{\textsc{reference} \\ \textsc{free}}}}
    & \cellcolor{highlight}\xcometensqe & \cellcolor{highlight}$0.83$ & \cellcolor{highlight}$80.40$ & \cellcolor{highlight}$83.47$ & \cellcolor{highlight}$81.40$ & \cellcolor{highlight}$0.92$ & \cellcolor{highlight}$70.00$ & \cellcolor{highlight}$63.60$ & \cellcolor{highlight}$67.73$ & \cellcolor{highlight}$41.40$ & \cellcolor{highlight}$-2.47$ \\
    & \cellcolor{highlight}\mbrmetricxqe & \cellcolor{highlight}$0.73$ & \cellcolor{highlight}$79.00$ & \cellcolor{highlight}$82.81$ & \cellcolor{highlight}$80.23$ & \cellcolor{highlight}$0.80$ & \cellcolor{highlight}$67.02$ & \cellcolor{highlight}$65.91$ & \cellcolor{highlight}$66.64$ & \cellcolor{highlight}$38.47$ & \cellcolor{highlight}$-2.40$ \\
    & \cellcolor{highlight}\metricxqe & \cellcolor{highlight}$-3.90$ & \cellcolor{highlight}$76.73$ & \cellcolor{highlight}$87.70$ & \cellcolor{highlight}$80.07$ & \cellcolor{highlight}$-1.31$ & \cellcolor{highlight}$67.76$ & \cellcolor{highlight}$67.85$ & \cellcolor{highlight}$67.79$ & \cellcolor{highlight}$37.55$ & \cellcolor{highlight}$-2.59$ \\
    & \metricxxlqe & $-3.57$ & $77.91$ & $83.36$ & $79.64$ & $-1.64$ & $67.15$ & $70.08$ & $68.10$ & $36.09$ & $-2.83$ \\
    & \cellcolor{highlight}\gemba & \cellcolor{highlight}$-5.00$ & \cellcolor{highlight}$82.41$ & \cellcolor{highlight}$79.99$ & \cellcolor{highlight}$81.59$ & \cellcolor{highlight}$-1.00$ & \cellcolor{highlight}$64.12$ & \cellcolor{highlight}$74.12$ & \cellcolor{highlight}$67.14$ & \cellcolor{highlight}$42.58$ & \cellcolor{highlight}$-2.30$ \\
    & \mateseqe & $-4.00$ & $73.72$ & $85.64$ & $77.30$ & $0.00$ & $55.43$ & $75.05$ & $60.72$ & $30.34$ & $-3.59$ \\
    & \cometqe & $-0.01$ & $75.35$ & $82.53$ & $77.60$ & $0.05$ & $59.64$ & $68.59$ & $62.35$ & $37.35$ & $-2.66$ \\
    & \cometmqmqe & $0.08$ & $75.40$ & $86.33$ & $78.72$ & $0.10$ & $61.63$ & $73.84$ & $65.22$ & $33.52$ & $-3.59$ \\
    & \cometkiwi & $0.76$ & $78.62$ & $80.90$ & $79.37$ & $0.80$ & $64.79$ & $66.52$ & $65.35$ & $39.28$ & $-2.61$ \\
    & \cometkiwixl & $0.64$ & $78.04$ & $79.81$ & $78.62$ & $0.71$ & $64.73$ & $65.51$ & $64.99$ & $38.78$ & $-2.60$ \\
    \cmidrule(lr){2-12}
    & \randombaseline & $-5.00$ & $64.06$ & $100.00$ & $72.78$ & $-4.00$ & $42.14$ & $99.99$ & $52.21$ & $29.04$ & $-3.74$ \\
    & \dasqm & $63.50$ & $67.83$ & $95.95$ & $75.18$ & $74.67$ & $48.30$ & $82.61$ & $56.06$ & $32.99$ & $-3.22$ \\
    \bottomrule
    \end{NiceTabular}
}
\caption[]{Metrics' Precision, Recall, and \fscore in binary classification, distinguishing \good from \bad, and \perfect from \other translations. $\tau$ is selected to maximize the \fscore \textbf{on the test set}. In the last two columns, we report metrics' Precision in translation re-ranking and the average MQM score of the selected translations. The test set is \wmtth and the translation direction is \langpair{zh}{en}. We report results concerning other translation directions in Appendix~\ref{apx:performance-results}. The metrics highlighted in grey are not openly available.\footnotemark}
\label{tab:performance-results-zhen}
\end{table*}

\begin{table*}[t]
\centering

\resizebox{0.8\linewidth}{!}{
    \begin{NiceTabular}{ll|rrrr|rrrr}[cell-space-limits=3pt]
    \toprule
     &  & \multicolumn{4}{c|}{\textbf{\good vs \bad}} & \multicolumn{4}{c}{\textbf{\perfect vs \other}} \\
    & \textbf{Metric} & \multicolumn{1}{c}{$\boldsymbol{\tau}$} & \multicolumn{1}{c}{\textbf{P}} & \multicolumn{1}{c}{\textbf{R}} & \multicolumn{1}{c|}{$\boldsymbol{F}$} & \multicolumn{1}{c}{$\boldsymbol{\tau}$} & \multicolumn{1}{c}{\textbf{P}} & \multicolumn{1}{c}{\textbf{R}} & \multicolumn{1}{c}{$\boldsymbol{F}$} \\
    \cmidrule(lr){2-10}
    \multirow{ 3 }{*}{\rotatebox{90}{\small \shortstack{\textsc{reference} \\ \textsc{based}}}}
    & \metricxxl & $-3.93$ & $76.58$ & $87.01$ & $79.77$ & $-2.97$ & $57.70$ & $88.80$ & $65.32$ \\
    & \comet & $0.77$ & $75.95$ & $75.28$ & $75.72$ & $0.79$ & $55.51$ & $74.57$ & $60.68$ \\
    & \bleurt & $0.61$ & $73.81$ & $79.92$ & $75.74$ & $0.64$ & $52.45$ & $76.89$ & $58.67$ \\
    \cmidrule(lr){2-10}
    \multirow{ 5 }{*}{\rotatebox{90}{\small \shortstack{\textsc{reference} \\ \textsc{free}}}}
    & \metricxxlqe & $-5.45$ & $73.36$ & $91.88$ & $78.64$ & $-3.54$ & $55.63$ & $90.32$ & $63.80$ \\
    & \cometmqmqe & $0.07$ & $72.57$ & $92.41$ & $78.17$ & $0.08$ & $52.59$ & $93.19$ & $61.53$ \\
    & \cometqe & $-0.02$ & $73.77$ & $85.81$ & $77.39$ & $-0.02$ & $50.54$ & $88.44$ & $58.96$ \\
    & \cometkiwi & $0.74$ & $75.57$ & $85.35$ & $78.58$ & $0.76$ & $53.38$ & $86.73$ & $61.23$ \\
    & \cometkiwixl & $0.62$ & $75.47$ & $83.37$ & $77.93$ & $0.64$ & $53.70$ & $85.03$ & $61.22$ \\
    \bottomrule
    \end{NiceTabular}
}

\caption{Metrics' Precision, Recall, and \fscore in binary classification, distinguish \good from \bad, and \perfect from \other translations. $\tau$ is selected to maximize the \fscore \textbf{on the development set}. The test set is \wmtth and the translation direction is \langpair{zh}{en}. Results in other translation directions are in Appendix~\ref{apx:performance-results}.}
\label{tab:performance-results-zhen-dev}
\end{table*}

\paragraph{Thresholds reliability}
Our results show that optimal thresholds tend to vary when moving from \wmtth to \wmttw for their calculation. For example, \metricxxlqe's $\tau$ shifts from \num{-3.57} to \num{-5.45}, and from \num{-1.64} to \num{-3.54}, when separating between \good and \bad and \perfect and \other, respectively, and other metrics display a similar pattern. Optimal threshold values do not appear stable across language pairs either, as illustrated in Figures~\ref{fig:thresholds-full-stability-4.0} and \ref{fig:thresholds-full-stability-1.0} in the Appendix. Specifically, optimal thresholds frequently differ between \langpair{en}{de} and the other translation directions in the \good vs \bad scenario, and are substantially lower for \langpair{he}{en} in the \perfect vs \other scenario. Such differences in the optimal thresholds might suggest that metric scores have different meanings depending on the translation direction. Furthermore, and as already discussed, an overly small optimal threshold might suggest that metrics are not precise enough. However, threshold values might also be influenced by the quality of the translations in the dataset. Indeed, on average, the \langpair{he}{en} dataset contains higher-quality translations compared to the other language pairs (Table~\ref{tab:mqm-statistics} in the Appendix). This might incentivize the metrics to ``settle'' for lower threshold values in order to maximize Recall. 

In general, we believe that the characteristics of the development set play an important role in determining appropriate thresholds for data filtering applications. Aside from the translation direction, evaluation datasets can differ in the MT systems employed to translate the source texts, the data domains included, and the human annotators who assessed translation quality. Therefore, while leaving the investigation of these phenomena to future work, we recommend estimating optimal thresholds using as much annotated data as possible, to prevent the peculiarities of any single dataset from overly influencing the estimated values. To support this, we are releasing our evaluation framework with options for estimating optimal metric thresholds across several datasets, depending on the user's Precision and Recall requirements.

\footnotetext{Due to its dependence on GPT-4, we include \gemba in the group of not openly available metrics.
Instead, openly-available versions of \metricx and \metricxqe can be found at \url{https://github.com/google-research/metricx}. However, we could not compute their predictions due to their high parameter count and thus resorted to using their outputs as submitted to WMT23, which were computed using different checkpoints than those currently available. Therefore, we include \metricx and \metricxqe in the group of not openly available metrics as well.}

\subsubsection{What is the human performance?} \label{seq:human-performance}
As a reference for human performance, we examine the agreement between two annotation schemas: DA+SQM and MQM. We use DA+SQM as a metric and show the results in the last row of Table~\ref{tab:performance-results-zhen}. It is important to note that DA+SQM annotations do not fully encompass the set of MQM annotations. Consequently, we evaluate its performance on a subset of the data, including $\approx 70\%$ of the data used for metrics, which means that their scores are not directly comparable. However, we observe that DA+SQM performs poorly in absolute terms, with particularly low Precision and a much higher Recall. We hypothesize that this might be due to DA+SQM's annotation guidelines, which instruct annotators to assign only a general quality score to translations rather than identifying specific translation errors, as is done within the MQM framework.\footnote{Within DA+SQM, human annotators rate a translation from $0$ to $100$ using a slider with $7$ marked levels, where each level is paired with a description of the corresponding translation quality.}  
This might lead to noisy annotations, rendering DA+SQM less suitable for fine-grained translation quality assessments. 

We further investigate this in Appendix~\ref{apx:dasqm-corr} by restricting the data available to MT metrics to that of DA+SQM, to estimate their segment-level correlations with MQM scores.
We find DA+SQM annotations correlate less strongly with MQM compared to all tested automatic metrics. 
We wish to emphasize that \citet{kocmi-etal-2023-findings} have already noted that DA+SQM-based annotations exhibit reduced precision in distinguishing between MT systems with similar performance, compared to MQM-based annotations. Furthermore, in a recent study, \citet{kocmi2024errorspanannotationbalanced} compared the correlation of several human annotation schemes with MQM and found that DA+SQM performed poorly. In line with these findings, our results raise additional concerns regarding DA+SQM reliability, showing performance inferior to automatic metrics.

\subsubsection{How \texorpdfstring{\bad}{bad} are false positives?}
\begin{figure}[t]
    \centering
    \resizebox{\columnwidth}{!}{
    \includegraphics{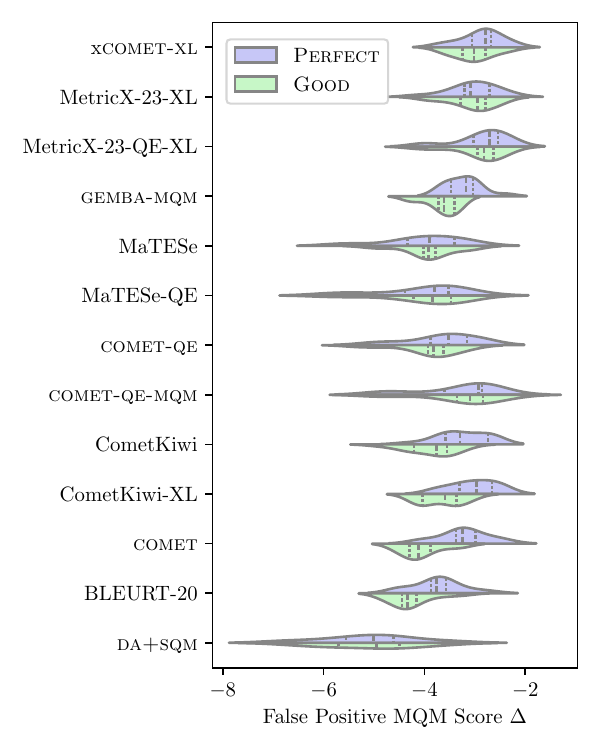}
    }
    \caption{Distribution of the MQM score $\Delta$ between the openly available metrics' false positive MQM scores and human thresholds, i.e., $-4$ for \good and $-1$ for \perfect. The dataset employed is the \langpair{zh}{en} split of \wmtth. Additional metrics are included in Figure~\ref{fig:fps-deltas-apx} in the Appendix.}
    \label{fig:fps-deltas-binary-wmt23-zhen}
\end{figure}

Our analysis suggests that MT metrics struggle to achieve high precision in binary classification. Concerning this, we are interested in assessing how \textit{bad} the false positives are -- i.e., translations that metrics mislabel as \good or \perfect. To this end, we plot in Figure~\ref{fig:fps-deltas-binary-wmt23-zhen} the distributions of the MQM score $\Delta$ computed for a false positive as the difference between the MQM score and the human threshold, which is $-4$ for a \good translation and $-1$ for a \perfect one.

The average false positive $\Delta$ ranges from $-4.25$ to $-2.85$ for both \good and \perfect classifications, indicating that the translations mislabeled by the best metrics contain an average of $\approx 3$ additional Minor errors. Overall, the  MQM $\Delta$ distributions of top-performing metrics are low-variance and skewed to the right, particularly when classifying \perfect translations. In contrast, less accurate metrics exhibit high-variance distributions, with the average $\Delta$ shifting towards lower values. Again, \dasqm performance is notably poor, showing the highest-variance distribution and the most leftward-shifted average.

\subsection{Translation Re-Ranking}
We present the results in the last two columns of Table~\ref{tab:performance-results-zhen}. To facilitate interpretation of these results, we also calculate the average MQM score of the translations ranked highest by MT metrics, and report it in the last column. Additionally, it is important to note that there are $15$ translations per source text.

As shown, metric precision ranges from $30\%$ to $43\%$, with the highest precision rates being $43.17\%$ and $42.58\%$, achieved by \xcometens and \gemba, respectively. Furthermore, the top-performing metrics yield average MQM scores of $\approx -2.50$, indicating that their highest-ranked translations contain an average of two and a half Minor errors. In contrast, human judgments suggest that the average MQM score of the highest-ranked translations is $-0.67$. 

Notably, reference-based metrics consistently outperform reference-free ones. By looking at pairs of metrics of the same family, \xcometens, \metricx, and \metricxxl outperform \xcometensqe, \metricxqe, and \metricxxlqe, respectively, in \langpair{zh}{en} and across the other translation directions (see Tables~\ref{tab:performance-results-ende-apx} and \ref{tab:performance-results-heen-apx} for results concerning \langpair{en}{de} and \langpair{he}{en}). However, in real-world translation re-ranking scenarios, references are not available. To account for this, we assess the performance of reference-based metrics when used as the utility function in an MBR decoding-like scenario, therefore not relying on the presence of reference translations.\footnote{MBR decoding seeks the candidate translation that maximizes an external notion of utility \cite{eikema-aziz-2022-sampling}. The set of candidate translations is used as both the set of hypotheses and to approximate a set of references. Then, each candidate translation is compared with all the others, employing reference-based metrics as the utility function.}

\paragraph{Metric performance in MBR decoding}
Table~\ref{tab:mbr} shows the translation re-ranking performance of reference-based metrics when used as the utility function for MBR decoding, and comparing it to the reference-based re-ranking scenario presented in the last columns of Tables~\ref{tab:performance-results-zhen}, \ref{tab:performance-results-ende-apx}, and \ref{tab:performance-results-heen-apx}. On average, the absence of reference translations reduces performance. However, this is not true for all language directions, as MBR decoding outperforms reference-based re-ranking in \langpair{zh}{en}. We believe this exception is due to the particularly poor quality of the references in the \langpair{zh}{en} split of \wmtth, as discussed by \citet{freitag-etal-2023-results}. 

\begin{table}[t]
    \centering
\resizebox{\columnwidth}{!}{
\begin{tabular}{lrrrrrr}
\toprule
& \multicolumn{2}{c}{\langpair{zh}{en}} &  \multicolumn{2}{c}{\langpair{en}{de}} &  \multicolumn{2}{c}{\langpair{he}{en}} \\ 
\textbf{Metric} & \textbf{MBR} & \textbf{Tab~\ref{tab:performance-results-zhen}} &  \textbf{MBR} & \textbf{Tab~\ref{tab:performance-results-ende-apx}} & \textbf{MBR} & \textbf{Tab~\ref{tab:performance-results-heen-apx}} \\ 
\midrule
\xcometxl & $40.27$ & $37.49$ & $44.03$ & $47.31$ & $65.05$ & $68.31$ \\
\metricxxl & $41.10$ & $39.52$ & $48.24$ & $47.81$ & $63.71$ & $67.17$ \\
\matese & $35.27$ & $33.07$ & $44.53$ & $43.18$ & $60.20$ & $61.99$ \\
\comet & $37.20$ & $34.25$ & $45.12$ & $48.26$ & $66.38$ & $70.01$ \\
\bleurt & $36.07$ & $33.35$ & $47.76$ & $48.27$ & $63.70$ & $68.33$ \\
\cmidrule(lr){2-7}
\#1 \textsc{ref free} & -- & $39.28$ & -- & $45.71$ & -- & $65.61$ \\
\#2 \textsc{ref free} & -- & $38.78$ & -- & $45.57$ & -- & $63.25$ \\
\bottomrule
\end{tabular}
}

    \caption{Re-Ranking Precision of reference-based metrics when used as the utility function for MBR decoding, compared with the reference-based re-ranking scenario in the last two columns of Tables~\ref{tab:performance-results-zhen}, \ref{tab:performance-results-ende-apx}, and \ref{tab:performance-results-heen-apx}. The last two rows of this table show the performance of the best and second-best reference-free metrics in translation re-ranking.}
    \label{tab:mbr}
\end{table} 

Furthermore, the last two rows of Table~\ref{tab:mbr} present the performance of the best and second-best openly available, reference-free metrics in translation re-ranking. 
Our results indicate that reference-based metrics used as the utility function for MBR decoding tend to outperform reference-free metrics. Specifically, the best performance is achieved by two reference-based metrics: \metricxxl, in \langpair{zh}{en} and \langpair{en}{de}, and \comet in \langpair{he}{en}. Again, by looking at metrics of the same family, \metricxxl outperforms \metricxxlqe across the board.
These findings suggest that translation re-ranking using MBR decoding may be more reliable than QE re-ranking.\footnote{This is based on the assumption that using translations generated by distinct MT systems can serve as an effective approximation of sampling from a single system. We delve into the differences between MBR decoding and our evaluation setup in the Limitations.} 
Previous studies have already compared MT metrics in MBR decoding and QE re-ranking. For example, \citet{freitag-etal-2022-high} and \citet{fernandes-etal-2022-quality} employ human annotators to evaluate the quality of translations produced by MT systems when using one or the other re-ranking technique. However, due to the high cost of human annotations, these studies were limited to including only a few metrics. Furthermore, their findings may need to be revisited to determine whether they remain valid with the introduction of new metrics. In contrast, 
by relying solely on the human annotations released annually at WMT, our setup facilitates updating results as soon as new metrics or datasets become available. 

\section{Related Work}
Previous studies have focused primarily on the problem of \textit{Error attribution}. Specifically, the Shared Task on Quality Estimation at WMT investigated the ability of MT metrics to predict word-level annotations \cite{zerva-etal-2022-findings, blain-etal-2023-findings}.
\citet{fomicheva-etal-2022-translation} and \citet{rei-etal-2023-inside} employed attribution methods to derive explanations for the predictions of MT metrics, measuring the faithfulness of such explanations by comparing them to human annotations. 
To tackle the same issue, 
\citet{perrella-etal-2022-matese}, \citet{fernandes-etal-2023-devil}, \citet{guerreiro2023xcomet}, \citet{kocmi-federmann-2023-gemba}, and \citet{xu-etal-2023-instructscore} proposed metrics that 
address the lack of \textit{Error attribution} by 
providing explanations in the form of either span-level annotations or natural language rationales. Furthermore, recent studies have introduced dedicated benchmarks to investigate the impact of specific translation errors, such as disambiguation errors \cite{campolungo-etal-2022-dibimt, 10.1162/coli_a_00541} and wrongly translated named entities \cite{conia-etal-2024-towards}.

In a different vein, and closer to our work, some studies explored the meaning of raw metrics scores in terms of their alignment with human judgments. \citet{mathur-etal-2020-tangled} studied the meaning of system-level score deltas for BLEU \cite{papineni-etal-2002-bleu}, showing that a statistically significant increase of $0$-$3$ BLEU points corresponds to significantly better MT systems less than half of the time, in terms of human judgments. Similarly, \citet{kocmi-etal-2024-navigating} investigated the relationship between MT metrics' system-level score deltas and human judgments. 
Finally, in a recent study, \citet{agrawal2024automaticmetricsassesshighquality} evaluated MT metrics' ability to assess high-quality translations by examining their correlation with human judgments, as well as their Precision, Recall, and \fscore, using a setup similar to ours. 
However, instead of calculating metrics thresholds from data, they arbitrarily assumed that a metric indicates \textit{high-quality} only if its normalized assessments fall within the $[0.99, 1.00]$ interval. In contrast, we measure metrics performance in data filtering without making assumptions about the meaning of their assessments, aiming to understand this meaning through the evaluation itself.

\section{Conclusion}
In this work, we introduce a novel evaluation framework for MT metrics. 
Within this framework, we measure metrics performance in i) binary classification, i.e., distinguishing between \good and \bad, and \perfect and \other translations, and ii) in a proxy scenario for translation re-ranking, selecting the best among the translations of the same source text. By measuring performance in terms of Precision, Recall, and $F$-score, we fulfill a dual purpose. First, we offer a more intuitive interpretation of metrics' capabilities, as compared to correlation with human judgment, and second, we provide concrete user recommendations concerning novel MT metric use cases. We find that MT metrics perform relatively well in distinguishing between \good and \bad translations, but struggle with Precision, especially when dealing with higher-quality translations like in the \perfect vs \other scenario. Our results show that \metricxxlqe is the best openly available metric for data filtering applications, while \metricxxl and \comet achieve the highest performance in translation re-ranking. Additionally, we demonstrate that reference-based MT metrics, when used as the utility function in an MBR decoding-like scenario, outperform reference-free ones, suggesting that MBR decoding may be superior to QE re-ranking.
Finally, we report notably poor performance for \dasqm annotations used as a metric within our evaluation framework, raising concerns about its reliability. 

\clearpage
\section{Limitations}

\paragraph{Language coverage}
We acknowledge that the scope of our work is limited by the available test data, covering only a few language directions. However, our evaluation framework is agnostic to the test data employed. Therefore, we leave the investigation of metric performance in more language directions to future works, depending on the availability of new annotated datasets.

\paragraph{Design choices in the data filtering scenario}
We made certain arbitrary decisions in the design of our framework and experimental setup. We chose $F_{\beta}$-score to select the optimal threshold $\tau$, with $\beta = \frac{1}{\sqrt{2}}$. While we explained our reasons for giving Precision a higher weight than Recall, it remains unclear whether $\beta = \frac{1}{\sqrt{2}}$ is the optimal choice. Furthermore, we selected the human score thresholds to be $-4$, for \good translations, and $-1$ for \perfect ones. We recognize that practitioners might have different requirements and may want to narrow or broaden these definitions. Therefore, we release our evaluation framework leaving this as an option for users. 

\paragraph{Evaluation fairness in the data filtering scenario}
In one of the two setups proposed, we selected the threshold $\tau$ to maximize the \fscore on the test set used for the evaluation. This optimization process might favor metrics whose assessments are more sensitive to the underlying gold score distribution, enabling them to achieve a better balance between Precision and Recall. As a result, discrete metrics -- i.e., those that output scores within a discrete set, such as the integers in $[-25, 0]$ for \gemba -- might be disadvantaged compared to continuous metrics -- i.e., those that output scores within a continuous interval, such as the real values in $[0,1]$ for metrics of the \comet family. However, we argue that this limitation is inherent to the nature of discrete metrics rather than a flaw in our evaluation framework. Indeed, studying the ability of MT metrics to distinguish between \good and \bad translations requires identifying the score threshold that best separates them, and discrete metrics inherently offer a much more limited set of options for optimizing this threshold. Nonetheless, if discrete metrics are indeed disadvantaged, using a development set could mitigate the impact of this phenomenon.

\paragraph{Alignment between the translation re-ranking scenario and the corresponding metric use cases}
We designed the translation re-ranking scenario as a proxy for QE re-ranking and MBR decoding. However, our setup differs from these two use cases in two ways:
\begin{enumerate}
    \item Candidates number: The test datasets we used feature $15$, $12$, and $13$ translations per source text, for \langpair{zh}{en}, \langpair{en}{de}, and \langpair{he}{en}, respectively. However, in QE re-ranking and MBR decoding it is common to work with a larger number of candidate translations, often reaching hundreds per source text.

    \item Candidates selection: In QE re-ranking and MBR decoding, candidate translations are typically sampled from the same MT system. In contrast, in our annotated datasets, each candidate translation was generated by a different MT system.
\end{enumerate}
In future work, it would be interesting to investigate whether our results might vary when dealing with a higher number of candidate translations or when all candidates are sampled from the same MT system.

\section*{Acknowledgements}

\begin{center}
\noindent
    \begin{minipage}{0.1\linewidth}
        \begin{center}
            \includegraphics[scale=0.05]{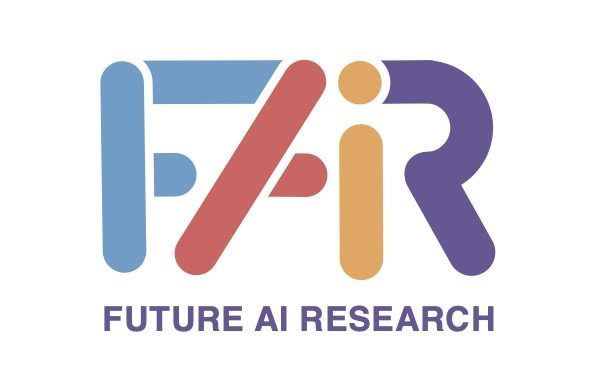}
        \end{center}
    \end{minipage}
    \hspace{0.01\linewidth}
    \begin{minipage}{0.70\linewidth}
         We gratefully acknowledge the support of the PNRR MUR project PE0000013-FAIR, and the CREATIVE project (CRoss-modal understanding and gEnerATIon of Visual and tExtual content), which is funded by the MUR Progetti di Rilevante Interesse Nazionale programme (PRIN 2020). 
    \end{minipage}
    \hspace{0.01\linewidth}
    \begin{minipage}{0.1\linewidth}
        \begin{center}
            \includegraphics[scale=0.08]{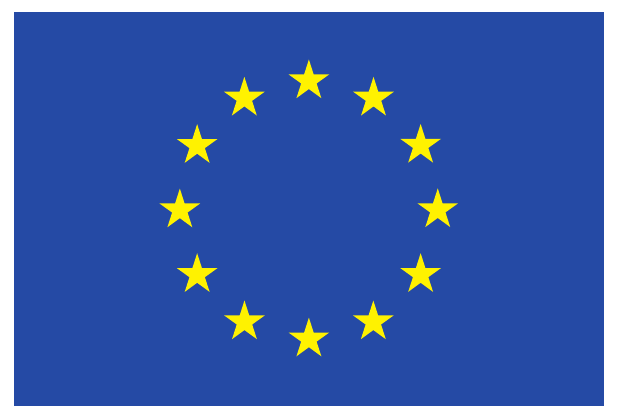}
        \end{center}
    \end{minipage}\\
\end{center}
\vspace{0.2cm}

\noindent This work was carried out while Lorenzo Proietti was enrolled in the Italian National Doctorate on Artificial Intelligence run by Sapienza University of Rome.

\bibliography{anthology,custom}

\begin{thebibliography}{74}
\providecommand{\natexlab}[1]{#1}

\bibitem[{Agrawal et~al.(2024)Agrawal, Farinhas, Rei, and Martins}]{agrawal2024automaticmetricsassesshighquality}
Sweta Agrawal, António Farinhas, Ricardo Rei, and André F.~T. Martins. 2024.
\newblock \href {https://arxiv.org/abs/2405.18348} {Can automatic metrics assess high-quality translations?}
\newblock \emph{arXiv preprint, arXiv:2405.18348}.

\bibitem[{Alves et~al.(2024)Alves, Pombal, Guerreiro, Martins, Alves, Farajian, Peters, Rei, Fernandes, Agrawal, Colombo, de~Souza, and Martins}]{alves2024tower}
Duarte~Miguel Alves, Jos{\'e} Pombal, Nuno~M Guerreiro, Pedro~Henrique Martins, Jo{\~a}o Alves, Amin Farajian, Ben Peters, Ricardo Rei, Patrick Fernandes, Sweta Agrawal, Pierre Colombo, Jos{\'e} G.~C. de~Souza, and Andre Martins. 2024.
\newblock \href {https://openreview.net/forum?id=EHPns3hVkj} {Tower: An open multilingual large language model for translation-related tasks}.
\newblock In \emph{Proceedings of the First Conference on Language Modeling}.

\bibitem[{{Barredo Arrieta} et~al.(2020){Barredo Arrieta}, Díaz-Rodríguez, {Del Ser}, Bennetot, Tabik, Barbado, Garcia, Gil-Lopez, Molina, Benjamins, Chatila, and Herrera}]{arrieta2019explainable}
Alejandro {Barredo Arrieta}, Natalia Díaz-Rodríguez, Javier {Del Ser}, Adrien Bennetot, Siham Tabik, Alberto Barbado, Salvador Garcia, Sergio Gil-Lopez, Daniel Molina, Richard Benjamins, Raja Chatila, and Francisco Herrera. 2020.
\newblock \href {https://doi.org/10.1016/j.inffus.2019.12.012} {Explainable artificial intelligence (xai): Concepts, taxonomies, opportunities and challenges toward responsible ai}.
\newblock \emph{Information Fusion}, 58:82--115.

\bibitem[{Blain et~al.(2023)Blain, Zerva, Rei, Guerreiro, Kanojia, C.~de Souza, Silva, Vaz, Jingxuan, Azadi, Orasan, and Martins}]{blain-etal-2023-findings}
Frederic Blain, Chrysoula Zerva, Ricardo Rei, Nuno~M. Guerreiro, Diptesh Kanojia, Jos{\'e}~G. C.~de Souza, Beatriz Silva, T{\^a}nia Vaz, Yan Jingxuan, Fatemeh Azadi, Constantin Orasan, and Andr{\'e} Martins. 2023.
\newblock \href {https://doi.org/10.18653/v1/2023.wmt-1.52} {Findings of the {WMT} 2023 shared task on quality estimation}.
\newblock In \emph{Proceedings of the Eighth Conference on Machine Translation}, pages 629--653, Singapore. Association for Computational Linguistics.

\bibitem[{Bojar et~al.(2017)Bojar, Graham, and Kamran}]{bojar-etal-2017-results}
Ond{\v{r}}ej Bojar, Yvette Graham, and Amir Kamran. 2017.
\newblock \href {https://doi.org/10.18653/v1/W17-4755} {Results of the {WMT}17 metrics shared task}.
\newblock In \emph{Proceedings of the Second Conference on Machine Translation}, pages 489--513, Copenhagen, Denmark. Association for Computational Linguistics.

\bibitem[{Bojar et~al.(2016)Bojar, Graham, Kamran, and Stanojevi{\'c}}]{bojar-etal-2016-results}
Ond{\v{r}}ej Bojar, Yvette Graham, Amir Kamran, and Milo{\v{s}} Stanojevi{\'c}. 2016.
\newblock \href {https://doi.org/10.18653/v1/W16-2302} {Results of the {WMT}16 metrics shared task}.
\newblock In \emph{Proceedings of the First Conference on Machine Translation: Volume 2, Shared Task Papers}, pages 199--231, Berlin, Germany. Association for Computational Linguistics.

\bibitem[{Campolungo et~al.(2022)Campolungo, Martelli, Saina, and Navigli}]{campolungo-etal-2022-dibimt}
Niccol{\`o} Campolungo, Federico Martelli, Francesco Saina, and Roberto Navigli. 2022.
\newblock \href {https://doi.org/10.18653/v1/2022.acl-long.298} {{D}i{B}i{MT}: A novel benchmark for measuring {W}ord {S}ense {D}isambiguation biases in {M}achine {T}ranslation}.
\newblock In \emph{Proceedings of the 60th Annual Meeting of the Association for Computational Linguistics (Volume 1: Long Papers)}, pages 4331--4352, Dublin, Ireland. Association for Computational Linguistics.

\bibitem[{Chi et~al.(2021)Chi, Dong, Wei, Yang, Singhal, Wang, Song, Mao, Huang, and Zhou}]{chi-etal-2021-infoxlm}
Zewen Chi, Li~Dong, Furu Wei, Nan Yang, Saksham Singhal, Wenhui Wang, Xia Song, Xian-Ling Mao, Heyan Huang, and Ming Zhou. 2021.
\newblock \href {https://doi.org/10.18653/v1/2021.naacl-main.280} {{I}nfo{XLM}: An information-theoretic framework for cross-lingual language model pre-training}.
\newblock In \emph{Proceedings of the 2021 Conference of the North American Chapter of the Association for Computational Linguistics: Human Language Technologies}, pages 3576--3588, Online. Association for Computational Linguistics.

\bibitem[{Chung et~al.(2021)Chung, Fevry, Tsai, Johnson, and Ruder}]{chung2021rethinking}
Hyung~Won Chung, Thibault Fevry, Henry Tsai, Melvin Johnson, and Sebastian Ruder. 2021.
\newblock \href {https://openreview.net/forum?id=xpFFI_NtgpW} {Rethinking embedding coupling in pre-trained language models}.
\newblock In \emph{Proceedings of the International Conference on Learning Representations}.

\bibitem[{Conia et~al.(2024)Conia, Lee, Li, Minhas, Potdar, and Li}]{conia-etal-2024-towards}
Simone Conia, Daniel Lee, Min Li, Umar~Farooq Minhas, Saloni Potdar, and Yunyao Li. 2024.
\newblock Towards cross-cultural machine translation with retrieval-augmented generation from multilingual knowledge graphs.
\newblock In \emph{Proceedings of the 2024 Conference on Empirical Methods in Natural Language Processing}, Miami, Florida, USA. Association for Computational Linguistics.

\bibitem[{Conneau et~al.(2020)Conneau, Khandelwal, Goyal, Chaudhary, Wenzek, Guzm{\'a}n, Grave, Ott, Zettlemoyer, and Stoyanov}]{conneau-etal-2020-unsupervised}
Alexis Conneau, Kartikay Khandelwal, Naman Goyal, Vishrav Chaudhary, Guillaume Wenzek, Francisco Guzm{\'a}n, Edouard Grave, Myle Ott, Luke Zettlemoyer, and Veselin Stoyanov. 2020.
\newblock \href {https://doi.org/10.18653/v1/2020.acl-main.747} {Unsupervised cross-lingual representation learning at scale}.
\newblock In \emph{Proceedings of the 58th Annual Meeting of the Association for Computational Linguistics}, pages 8440--8451, Online. Association for Computational Linguistics.

\bibitem[{Deutsch et~al.(2023)Deutsch, Foster, and Freitag}]{deutsch-etal-2023-ties}
Daniel Deutsch, George Foster, and Markus Freitag. 2023.
\newblock \href {https://doi.org/10.18653/v1/2023.emnlp-main.798} {Ties matter: Meta-evaluating modern metrics with pairwise accuracy and tie calibration}.
\newblock In \emph{Proceedings of the 2023 Conference on Empirical Methods in Natural Language Processing}, pages 12914--12929, Singapore. Association for Computational Linguistics.

\bibitem[{Dreano et~al.(2023)Dreano, Molloy, and Murphy}]{dreano-etal-2023-tokengram}
S{\"o}ren Dreano, Derek Molloy, and Noel Murphy. 2023.
\newblock \href {https://doi.org/10.18653/v1/2023.wmt-1.59} {{T}okengram{\_}{F}, a fast and accurate token-based chr{F}++ derivative}.
\newblock In \emph{Proceedings of the Eighth Conference on Machine Translation}, pages 730--737, Singapore. Association for Computational Linguistics.

\bibitem[{Eikema and Aziz(2020)}]{eikema-aziz-2020-map}
Bryan Eikema and Wilker Aziz. 2020.
\newblock \href {https://doi.org/10.18653/v1/2020.coling-main.398} {Is {MAP} decoding all you need? the inadequacy of the mode in neural machine translation}.
\newblock In \emph{Proceedings of the 28th International Conference on Computational Linguistics}, pages 4506--4520, Barcelona, Spain (Online). International Committee on Computational Linguistics.

\bibitem[{Eikema and Aziz(2022)}]{eikema-aziz-2022-sampling}
Bryan Eikema and Wilker Aziz. 2022.
\newblock \href {https://doi.org/10.18653/v1/2022.emnlp-main.754} {Sampling-based approximations to minimum {B}ayes risk decoding for neural machine translation}.
\newblock In \emph{Proceedings of the 2022 Conference on Empirical Methods in Natural Language Processing}, pages 10978--10993, Abu Dhabi, United Arab Emirates. Association for Computational Linguistics.

\bibitem[{ElNokrashy and Kocmi(2023)}]{elnokrashy-kocmi-2023-ebleu}
Muhammad ElNokrashy and Tom Kocmi. 2023.
\newblock \href {https://doi.org/10.18653/v1/2023.wmt-1.61} {e{BLEU}: Unexpectedly good machine translation evaluation using simple word embeddings}.
\newblock In \emph{Proceedings of the Eighth Conference on Machine Translation}, pages 746--750, Singapore. Association for Computational Linguistics.

\bibitem[{Farinhas et~al.(2023)Farinhas, de~Souza, and Martins}]{farinhas-etal-2023-empirical}
Ant{\'o}nio Farinhas, Jos{\'e} de~Souza, and Andre Martins. 2023.
\newblock \href {https://doi.org/10.18653/v1/2023.emnlp-main.733} {An empirical study of translation hypothesis ensembling with large language models}.
\newblock In \emph{Proceedings of the 2023 Conference on Empirical Methods in Natural Language Processing}, pages 11956--11970, Singapore. Association for Computational Linguistics.

\bibitem[{Fernandes et~al.(2023)Fernandes, Deutsch, Finkelstein, Riley, Martins, Neubig, Garg, Clark, Freitag, and Firat}]{fernandes-etal-2023-devil}
Patrick Fernandes, Daniel Deutsch, Mara Finkelstein, Parker Riley, Andr{\'e} Martins, Graham Neubig, Ankush Garg, Jonathan Clark, Markus Freitag, and Orhan Firat. 2023.
\newblock \href {https://doi.org/10.18653/v1/2023.wmt-1.100} {The devil is in the errors: Leveraging large language models for fine-grained machine translation evaluation}.
\newblock In \emph{Proceedings of the Eighth Conference on Machine Translation}, pages 1066--1083, Singapore. Association for Computational Linguistics.

\bibitem[{Fernandes et~al.(2022)Fernandes, Farinhas, Rei, C.~de Souza, Ogayo, Neubig, and Martins}]{fernandes-etal-2022-quality}
Patrick Fernandes, Ant{\'o}nio Farinhas, Ricardo Rei, Jos{\'e}~G. C.~de Souza, Perez Ogayo, Graham Neubig, and Andre Martins. 2022.
\newblock \href {https://doi.org/10.18653/v1/2022.naacl-main.100} {Quality-aware decoding for neural machine translation}.
\newblock In \emph{Proceedings of the 2022 Conference of the North American Chapter of the Association for Computational Linguistics: Human Language Technologies}, pages 1396--1412, Seattle, United States. Association for Computational Linguistics.

\bibitem[{Finkelstein and Freitag(2024)}]{finkelstein2024mbr}
Mara Finkelstein and Markus Freitag. 2024.
\newblock \href {https://openreview.net/forum?id=bkNx3O0sND} {{MBR} and {QE} finetuning: Training-time distillation of the best and most expensive decoding methods}.
\newblock In \emph{Proceedings of The Twelfth International Conference on Learning Representations}.

\bibitem[{Fomicheva et~al.(2022{\natexlab{a}})Fomicheva, Specia, and Aletras}]{fomicheva-etal-2022-translation}
Marina Fomicheva, Lucia Specia, and Nikolaos Aletras. 2022{\natexlab{a}}.
\newblock \href {https://doi.org/10.18653/v1/2022.findings-acl.327} {Translation error detection as rationale extraction}.
\newblock In \emph{Findings of the Association for Computational Linguistics: ACL 2022}, pages 4148--4159, Dublin, Ireland. Association for Computational Linguistics.

\bibitem[{Fomicheva et~al.(2022{\natexlab{b}})Fomicheva, Sun, Fonseca, Zerva, Blain, Chaudhary, Guzm{\'a}n, Lopatina, Specia, and Martins}]{fomicheva-etal-2022-mlqe}
Marina Fomicheva, Shuo Sun, Erick Fonseca, Chrysoula Zerva, Fr{\'e}d{\'e}ric Blain, Vishrav Chaudhary, Francisco Guzm{\'a}n, Nina Lopatina, Lucia Specia, and Andr{\'e} F.~T. Martins. 2022{\natexlab{b}}.
\newblock \href {https://aclanthology.org/2022.lrec-1.530} {{MLQE}-{PE}: A multilingual quality estimation and post-editing dataset}.
\newblock In \emph{Proceedings of the Thirteenth Language Resources and Evaluation Conference}, pages 4963--4974, Marseille, France. European Language Resources Association.

\bibitem[{Freitag et~al.(2021{\natexlab{a}})Freitag, Foster, Grangier, Ratnakar, Tan, and Macherey}]{freitag-etal-2021-experts}
Markus Freitag, George Foster, David Grangier, Viresh Ratnakar, Qijun Tan, and Wolfgang Macherey. 2021{\natexlab{a}}.
\newblock \href {https://doi.org/10.1162/tacl_a_00437} {Experts, errors, and context: A large-scale study of human evaluation for machine translation}.
\newblock \emph{Transactions of the Association for Computational Linguistics}, 9:1460--1474.

\bibitem[{Freitag et~al.(2022{\natexlab{a}})Freitag, Grangier, Tan, and Liang}]{freitag-etal-2022-high}
Markus Freitag, David Grangier, Qijun Tan, and Bowen Liang. 2022{\natexlab{a}}.
\newblock \href {https://doi.org/10.1162/tacl_a_00491} {High quality rather than high model probability: Minimum {B}ayes risk decoding with neural metrics}.
\newblock \emph{Transactions of the Association for Computational Linguistics}, 10:811--825.

\bibitem[{Freitag et~al.(2023)Freitag, Mathur, Lo, Avramidis, Rei, Thompson, Kocmi, Blain, Deutsch, Stewart, Zerva, Castilho, Lavie, and Foster}]{freitag-etal-2023-results}
Markus Freitag, Nitika Mathur, Chi-kiu Lo, Eleftherios Avramidis, Ricardo Rei, Brian Thompson, Tom Kocmi, Frederic Blain, Daniel Deutsch, Craig Stewart, Chrysoula Zerva, Sheila Castilho, Alon Lavie, and George Foster. 2023.
\newblock \href {https://doi.org/10.18653/v1/2023.wmt-1.51} {Results of {WMT}23 metrics shared task: Metrics might be guilty but references are not innocent}.
\newblock In \emph{Proceedings of the Eighth Conference on Machine Translation}, pages 578--628, Singapore. Association for Computational Linguistics.

\bibitem[{Freitag et~al.(2022{\natexlab{b}})Freitag, Rei, Mathur, Lo, Stewart, Avramidis, Kocmi, Foster, Lavie, and Martins}]{freitag-etal-2022-results}
Markus Freitag, Ricardo Rei, Nitika Mathur, Chi-kiu Lo, Craig Stewart, Eleftherios Avramidis, Tom Kocmi, George Foster, Alon Lavie, and Andr{\'e} F.~T. Martins. 2022{\natexlab{b}}.
\newblock \href {https://aclanthology.org/2022.wmt-1.2} {Results of {WMT}22 metrics shared task: Stop using {BLEU} {--} neural metrics are better and more robust}.
\newblock In \emph{Proceedings of the Seventh Conference on Machine Translation (WMT)}, pages 46--68, Abu Dhabi, United Arab Emirates (Hybrid). Association for Computational Linguistics.

\bibitem[{Freitag et~al.(2021{\natexlab{b}})Freitag, Rei, Mathur, Lo, Stewart, Foster, Lavie, and Bojar}]{freitag-etal-2021-results}
Markus Freitag, Ricardo Rei, Nitika Mathur, Chi-kiu Lo, Craig Stewart, George Foster, Alon Lavie, and Ond{\v{r}}ej Bojar. 2021{\natexlab{b}}.
\newblock \href {https://aclanthology.org/2021.wmt-1.73} {Results of the {WMT}21 metrics shared task: Evaluating metrics with expert-based human evaluations on {TED} and news domain}.
\newblock In \emph{Proceedings of the Sixth Conference on Machine Translation}, pages 733--774, Online. Association for Computational Linguistics.

\bibitem[{Goyal et~al.(2021)Goyal, Du, Ott, Anantharaman, and Conneau}]{goyal-etal-2021-larger}
Naman Goyal, Jingfei Du, Myle Ott, Giri Anantharaman, and Alexis Conneau. 2021.
\newblock \href {https://doi.org/10.18653/v1/2021.repl4nlp-1.4} {Larger-scale transformers for multilingual masked language modeling}.
\newblock In \emph{Proceedings of the 6th Workshop on Representation Learning for NLP (RepL4NLP-2021)}, pages 29--33, Online. Association for Computational Linguistics.

\bibitem[{Goyal et~al.(2022)Goyal, Gao, Chaudhary, Chen, Wenzek, Ju, Krishnan, Ranzato, Guzm{\'a}n, and Fan}]{goyal-etal-2022-flores}
Naman Goyal, Cynthia Gao, Vishrav Chaudhary, Peng-Jen Chen, Guillaume Wenzek, Da~Ju, Sanjana Krishnan, Marc{'}Aurelio Ranzato, Francisco Guzm{\'a}n, and Angela Fan. 2022.
\newblock \href {https://doi.org/10.1162/tacl_a_00474} {The {F}lores-101 evaluation benchmark for low-resource and multilingual machine translation}.
\newblock \emph{Transactions of the Association for Computational Linguistics}, 10:522--538.

\bibitem[{Graham et~al.(2013)Graham, Baldwin, Moffat, and Zobel}]{graham-etal-2013-continuous}
Yvette Graham, Timothy Baldwin, Alistair Moffat, and Justin Zobel. 2013.
\newblock \href {https://aclanthology.org/W13-2305} {Continuous measurement scales in human evaluation of machine translation}.
\newblock In \emph{Proceedings of the 7th Linguistic Annotation Workshop and Interoperability with Discourse}, pages 33--41, Sofia, Bulgaria. Association for Computational Linguistics.

\bibitem[{Guerreiro et~al.(2024)Guerreiro, Rei, Stigt, Coheur, Colombo, and Martins}]{guerreiro2023xcomet}
Nuno~M. Guerreiro, Ricardo Rei, Daan~van Stigt, Luisa Coheur, Pierre Colombo, and André F.~T. Martins. 2024.
\newblock \href {https://doi.org/10.1162/tacl_a_00683} {{xcomet: Transparent Machine Translation Evaluation through Fine-grained Error Detection}}.
\newblock \emph{Transactions of the Association for Computational Linguistics}, 12:979--995.

\bibitem[{Gulcehre et~al.(2023)Gulcehre, Paine, Srinivasan, Konyushkova, Weerts, Sharma, Siddhant, Ahern, Wang, Gu, Macherey, Doucet, Firat, and de~Freitas}]{gulcehre2023reinforced}
Caglar Gulcehre, Tom~Le Paine, Srivatsan Srinivasan, Ksenia Konyushkova, Lotte Weerts, Abhishek Sharma, Aditya Siddhant, Alex Ahern, Miaosen Wang, Chenjie Gu, Wolfgang Macherey, Arnaud Doucet, Orhan Firat, and Nando de~Freitas. 2023.
\newblock \href {https://arxiv.org/abs/2308.08998} {Reinforced self-training (rest) for language modeling}.
\newblock \emph{arXiv preprint, arXiv:2308.08998}.

\bibitem[{He et~al.(2023)He, Gao, and Chen}]{he2023debertav3}
Pengcheng He, Jianfeng Gao, and Weizhu Chen. 2023.
\newblock \href {https://openreview.net/forum?id=sE7-XhLxHA} {De{BERT}av3: Improving de{BERT}a using {ELECTRA}-style pre-training with gradient-disentangled embedding sharing}.
\newblock In \emph{Proceedings of The Eleventh International Conference on Learning Representations}.

\bibitem[{He et~al.(2024)He, Wang, Jiao, Zhang, Wang, Shi, and Tu}]{he-etal-2024-improving}
Zhiwei He, Xing Wang, Wenxiang Jiao, Zhuosheng Zhang, Rui Wang, Shuming Shi, and Zhaopeng Tu. 2024.
\newblock \href {https://doi.org/10.18653/v1/2024.naacl-long.451} {Improving machine translation with human feedback: An exploration of quality estimation as a reward model}.
\newblock In \emph{Proceedings of the 2024 Conference of the North American Chapter of the Association for Computational Linguistics: Human Language Technologies (Volume 1: Long Papers)}, pages 8164--8180, Mexico City, Mexico. Association for Computational Linguistics.

\bibitem[{Juraska et~al.(2023)Juraska, Finkelstein, Deutsch, Siddhant, Mirzazadeh, and Freitag}]{juraska-etal-2023-metricx}
Juraj Juraska, Mara Finkelstein, Daniel Deutsch, Aditya Siddhant, Mehdi Mirzazadeh, and Markus Freitag. 2023.
\newblock \href {https://doi.org/10.18653/v1/2023.wmt-1.63} {{M}etric{X}-23: The {G}oogle submission to the {WMT} 2023 metrics shared task}.
\newblock In \emph{Proceedings of the Eighth Conference on Machine Translation}, pages 756--767, Singapore. Association for Computational Linguistics.

\bibitem[{Karpinska et~al.(2022)Karpinska, Raj, Thai, Song, Gupta, and Iyyer}]{karpinska-etal-2022-demetr}
Marzena Karpinska, Nishant Raj, Katherine Thai, Yixiao Song, Ankita Gupta, and Mohit Iyyer. 2022.
\newblock \href {https://doi.org/10.18653/v1/2022.emnlp-main.649} {{DEMETR}: Diagnosing evaluation metrics for translation}.
\newblock In \emph{Proceedings of the 2022 Conference on Empirical Methods in Natural Language Processing}, pages 9540--9561, Abu Dhabi, United Arab Emirates. Association for Computational Linguistics.

\bibitem[{Kocmi et~al.(2023)Kocmi, Avramidis, Bawden, Bojar, Dvorkovich, Federmann, Fishel, Freitag, Gowda, Grundkiewicz, Haddow, Koehn, Marie, Monz, Morishita, Murray, Nagata, Nakazawa, Popel, Popovi{\'c}, and Shmatova}]{kocmi-etal-2023-findings}
Tom Kocmi, Eleftherios Avramidis, Rachel Bawden, Ond{\v{r}}ej Bojar, Anton Dvorkovich, Christian Federmann, Mark Fishel, Markus Freitag, Thamme Gowda, Roman Grundkiewicz, Barry Haddow, Philipp Koehn, Benjamin Marie, Christof Monz, Makoto Morishita, Kenton Murray, Makoto Nagata, Toshiaki Nakazawa, Martin Popel, Maja Popovi{\'c}, and Mariya Shmatova. 2023.
\newblock \href {https://doi.org/10.18653/v1/2023.wmt-1.1} {Findings of the 2023 conference on machine translation ({WMT}23): {LLM}s are here but not quite there yet}.
\newblock In \emph{Proceedings of the Eighth Conference on Machine Translation}, pages 1--42, Singapore. Association for Computational Linguistics.

\bibitem[{Kocmi et~al.(2022)Kocmi, Bawden, Bojar, Dvorkovich, Federmann, Fishel, Gowda, Graham, Grundkiewicz, Haddow, Knowles, Koehn, Monz, Morishita, Nagata, Nakazawa, Nov{\'a}k, Popel, and Popovi{\'c}}]{kocmi-etal-2022-findings}
Tom Kocmi, Rachel Bawden, Ond{\v{r}}ej Bojar, Anton Dvorkovich, Christian Federmann, Mark Fishel, Thamme Gowda, Yvette Graham, Roman Grundkiewicz, Barry Haddow, Rebecca Knowles, Philipp Koehn, Christof Monz, Makoto Morishita, Masaaki Nagata, Toshiaki Nakazawa, Michal Nov{\'a}k, Martin Popel, and Maja Popovi{\'c}. 2022.
\newblock \href {https://aclanthology.org/2022.wmt-1.1} {Findings of the 2022 conference on machine translation ({WMT}22)}.
\newblock In \emph{Proceedings of the Seventh Conference on Machine Translation (WMT)}, pages 1--45, Abu Dhabi, United Arab Emirates (Hybrid). Association for Computational Linguistics.

\bibitem[{Kocmi and Federmann(2023)}]{kocmi-federmann-2023-gemba}
Tom Kocmi and Christian Federmann. 2023.
\newblock \href {https://doi.org/10.18653/v1/2023.wmt-1.64} {{GEMBA}-{MQM}: Detecting translation quality error spans with {GPT}-4}.
\newblock In \emph{Proceedings of the Eighth Conference on Machine Translation}, pages 768--775, Singapore. Association for Computational Linguistics.

\bibitem[{Kocmi et~al.(2024{\natexlab{a}})Kocmi, Zouhar, Federmann, and Post}]{kocmi-etal-2024-navigating}
Tom Kocmi, Vil{\'e}m Zouhar, Christian Federmann, and Matt Post. 2024{\natexlab{a}}.
\newblock \href {https://doi.org/10.18653/v1/2024.acl-long.110} {Navigating the metrics maze: Reconciling score magnitudes and accuracies}.
\newblock In \emph{Proceedings of the 62nd Annual Meeting of the Association for Computational Linguistics (Volume 1: Long Papers)}, pages 1999--2014, Bangkok, Thailand. Association for Computational Linguistics.

\bibitem[{Kocmi et~al.(2024{\natexlab{b}})Kocmi, Zouhar, Avramidis, Grundkiewicz, Karpinska, Popović, Sachan, and Shmatova}]{kocmi2024errorspanannotationbalanced}
Tom Kocmi, Vilém Zouhar, Eleftherios Avramidis, Roman Grundkiewicz, Marzena Karpinska, Maja Popović, Mrinmaya Sachan, and Mariya Shmatova. 2024{\natexlab{b}}.
\newblock \href {https://arxiv.org/abs/2406.11580} {Error span annotation: A balanced approach for human evaluation of machine translation}.
\newblock \emph{arXiv preprint, arXiv:2406.11580}.

\bibitem[{Kudo(2018)}]{kudo-2018-subword}
Taku Kudo. 2018.
\newblock \href {https://doi.org/10.18653/v1/P18-1007} {Subword regularization: Improving neural network translation models with multiple subword candidates}.
\newblock In \emph{Proceedings of the 56th Annual Meeting of the Association for Computational Linguistics (Volume 1: Long Papers)}, pages 66--75, Melbourne, Australia. Association for Computational Linguistics.

\bibitem[{Kumar and Byrne(2004)}]{kumar-byrne-2004-minimum}
Shankar Kumar and William Byrne. 2004.
\newblock \href {https://aclanthology.org/N04-1022} {Minimum {B}ayes-risk decoding for statistical machine translation}.
\newblock In \emph{Proceedings of the Human Language Technology Conference of the North {A}merican Chapter of the Association for Computational Linguistics: {HLT}-{NAACL} 2004}, pages 169--176, Boston, Massachusetts, USA. Association for Computational Linguistics.

\bibitem[{Lommel et~al.(2014)Lommel, Burchardt, and Uszkoreit}]{mqm-framework}
Arle Lommel, Aljoscha Burchardt, and Hans Uszkoreit. 2014.
\newblock \href {https://doi.org/10.5565/rev/tradumatica.77} {Multidimensional quality metrics (mqm): A framework for declaring and describing translation quality metrics}.
\newblock \emph{Tradumàtica: tecnologies de la traducció}, 0:455--463.

\bibitem[{Ma et~al.(2018)Ma, Bojar, and Graham}]{ma-etal-2018-results}
Qingsong Ma, Ond{\v{r}}ej Bojar, and Yvette Graham. 2018.
\newblock \href {https://doi.org/10.18653/v1/W18-6450} {Results of the {WMT}18 metrics shared task: Both characters and embeddings achieve good performance}.
\newblock In \emph{Proceedings of the Third Conference on Machine Translation: Shared Task Papers}, pages 671--688, Belgium, Brussels. Association for Computational Linguistics.

\bibitem[{Ma et~al.(2019)Ma, Wei, Bojar, and Graham}]{ma-etal-2019-results}
Qingsong Ma, Johnny Wei, Ond{\v{r}}ej Bojar, and Yvette Graham. 2019.
\newblock \href {https://doi.org/10.18653/v1/W19-5302} {Results of the {WMT}19 metrics shared task: Segment-level and strong {MT} systems pose big challenges}.
\newblock In \emph{Proceedings of the Fourth Conference on Machine Translation (Volume 2: Shared Task Papers, Day 1)}, pages 62--90, Florence, Italy. Association for Computational Linguistics.

\bibitem[{Martelli et~al.(2024)Martelli, Perrella, Campolungo, Munda, Koeva, Tiberius, and Navigli}]{10.1162/coli_a_00541}
Federico Martelli, Stefano Perrella, Niccolò Campolungo, Tina Munda, Svetla Koeva, Carole Tiberius, and Roberto Navigli. 2024.
\newblock \href {https://doi.org/10.1162/coli_a_00541} {{DiBiMT: A Gold Evaluation Benchmark for Studying Lexical Ambiguity in Machine Translation}}.
\newblock \emph{Computational Linguistics}, pages 1--79.

\bibitem[{Mathur et~al.(2020{\natexlab{a}})Mathur, Baldwin, and Cohn}]{mathur-etal-2020-tangled}
Nitika Mathur, Timothy Baldwin, and Trevor Cohn. 2020{\natexlab{a}}.
\newblock \href {https://doi.org/10.18653/v1/2020.acl-main.448} {Tangled up in {BLEU}: Reevaluating the evaluation of automatic machine translation evaluation metrics}.
\newblock In \emph{Proceedings of the 58th Annual Meeting of the Association for Computational Linguistics}, pages 4984--4997, Online. Association for Computational Linguistics.

\bibitem[{Mathur et~al.(2020{\natexlab{b}})Mathur, Wei, Freitag, Ma, and Bojar}]{mathur-etal-2020-results}
Nitika Mathur, Johnny Wei, Markus Freitag, Qingsong Ma, and Ond{\v{r}}ej Bojar. 2020{\natexlab{b}}.
\newblock \href {https://aclanthology.org/2020.wmt-1.77} {Results of the {WMT}20 metrics shared task}.
\newblock In \emph{Proceedings of the Fifth Conference on Machine Translation}, pages 688--725, Online. Association for Computational Linguistics.

\bibitem[{Naskar et~al.(2023)Naskar, Deutsch, and Freitag}]{naskar-etal-2023-quality}
Subhajit Naskar, Daniel Deutsch, and Markus Freitag. 2023.
\newblock \href {https://doi.org/10.18653/v1/2023.wmt-1.67} {Quality estimation using minimum {B}ayes risk}.
\newblock In \emph{Proceedings of the Eighth Conference on Machine Translation}, pages 806--811, Singapore. Association for Computational Linguistics.

\bibitem[{OpenAI et~al.(2024)OpenAI, Achiam, Adler, Agarwal, Ahmad, Akkaya, Aleman, Almeida, Altenschmidt, Altman, Anadkat, Avila, Babuschkin, Balaji, Balcom, Baltescu, Bao, Bavarian, Belgum, Bello, Berdine, Bernadett-Shapiro, Berner, Bogdonoff, Boiko, Boyd, Brakman, Brockman, Brooks, Brundage, Button, Cai, Campbell, Cann, Carey, Carlson, Carmichael, Chan, Chang, Chantzis, Chen, Chen, Chen, Chen, Chen, Chess, Cho, Chu, Chung, Cummings, Currier, Dai, Decareaux, Degry, Deutsch, Deville, Dhar, Dohan, Dowling, Dunning, Ecoffet, Eleti, Eloundou, Farhi, Fedus, Felix, Fishman, Forte, Fulford, Gao, Georges, Gibson, Goel, Gogineni, Goh, Gontijo-Lopes, Gordon, Grafstein, Gray, Greene, Gross, Gu, Guo, Hallacy, Han, Harris, He, Heaton, Heidecke, Hesse, Hickey, Hickey, Hoeschele, Houghton, Hsu, Hu, Hu, Huizinga, Jain, Jain, Jang, Jiang, Jiang, Jin, Jin, Jomoto, Jonn, Jun, Kaftan, Łukasz Kaiser, Kamali, Kanitscheider, Keskar, Khan, Kilpatrick, Kim, Kim, Kim, Kirchner, Kiros, Knight, Kokotajlo, Łukasz Kondraciuk,
  Kondrich, Konstantinidis, Kosic, Krueger, Kuo, Lampe, Lan, Lee, Leike, Leung, Levy, Li, Lim, Lin, Lin, Litwin, Lopez, Lowe, Lue, Makanju, Malfacini, Manning, Markov, Markovski, Martin, Mayer, Mayne, McGrew, McKinney, McLeavey, McMillan, McNeil, Medina, Mehta, Menick, Metz, Mishchenko, Mishkin, Monaco, Morikawa, Mossing, Mu, Murati, Murk, Mély, Nair, Nakano, Nayak, Neelakantan, Ngo, Noh, Ouyang, O'Keefe, Pachocki, Paino, Palermo, Pantuliano, Parascandolo, Parish, Parparita, Passos, Pavlov, Peng, Perelman, de~Avila Belbute~Peres, Petrov, de~Oliveira~Pinto, Michael, Pokorny, Pokrass, Pong, Powell, Power, Power, Proehl, Puri, Radford, Rae, Ramesh, Raymond, Real, Rimbach, Ross, Rotsted, Roussez, Ryder, Saltarelli, Sanders, Santurkar, Sastry, Schmidt, Schnurr, Schulman, Selsam, Sheppard, Sherbakov, Shieh, Shoker, Shyam, Sidor, Sigler, Simens, Sitkin, Slama, Sohl, Sokolowsky, Song, Staudacher, Such, Summers, Sutskever, Tang, Tezak, Thompson, Tillet, Tootoonchian, Tseng, Tuggle, Turley, Tworek, Uribe, Vallone,
  Vijayvergiya, Voss, Wainwright, Wang, Wang, Wang, Ward, Wei, Weinmann, Welihinda, Welinder, Weng, Weng, Wiethoff, Willner, Winter, Wolrich, Wong, Workman, Wu, Wu, Wu, Xiao, Xu, Yoo, Yu, Yuan, Zaremba, Zellers, Zhang, Zhang, Zhao, Zheng, Zhuang, Zhuk, and Zoph}]{openai2024gpt4}
OpenAI, Josh Achiam, Steven Adler, Sandhini Agarwal, Lama Ahmad, Ilge Akkaya, Florencia~Leoni Aleman, Diogo Almeida, Janko Altenschmidt, Sam Altman, Shyamal Anadkat, Red Avila, Igor Babuschkin, Suchir Balaji, Valerie Balcom, Paul Baltescu, Haiming Bao, Mohammad Bavarian, Jeff Belgum, Irwan Bello, Jake Berdine, Gabriel Bernadett-Shapiro, Christopher Berner, Lenny Bogdonoff, Oleg Boiko, Madelaine Boyd, Anna-Luisa Brakman, Greg Brockman, Tim Brooks, Miles Brundage, Kevin Button, Trevor Cai, Rosie Campbell, Andrew Cann, Brittany Carey, Chelsea Carlson, Rory Carmichael, Brooke Chan, Che Chang, Fotis Chantzis, Derek Chen, Sully Chen, Ruby Chen, Jason Chen, Mark Chen, Ben Chess, Chester Cho, Casey Chu, Hyung~Won Chung, Dave Cummings, Jeremiah Currier, Yunxing Dai, Cory Decareaux, Thomas Degry, Noah Deutsch, Damien Deville, Arka Dhar, David Dohan, Steve Dowling, Sheila Dunning, Adrien Ecoffet, Atty Eleti, Tyna Eloundou, David Farhi, Liam Fedus, Niko Felix, Simón~Posada Fishman, Juston Forte, Isabella Fulford, Leo
  Gao, Elie Georges, Christian Gibson, Vik Goel, Tarun Gogineni, Gabriel Goh, Rapha Gontijo-Lopes, Jonathan Gordon, Morgan Grafstein, Scott Gray, Ryan Greene, Joshua Gross, Shixiang~Shane Gu, Yufei Guo, Chris Hallacy, Jesse Han, Jeff Harris, Yuchen He, Mike Heaton, Johannes Heidecke, Chris Hesse, Alan Hickey, Wade Hickey, Peter Hoeschele, Brandon Houghton, Kenny Hsu, Shengli Hu, Xin Hu, Joost Huizinga, Shantanu Jain, Shawn Jain, Joanne Jang, Angela Jiang, Roger Jiang, Haozhun Jin, Denny Jin, Shino Jomoto, Billie Jonn, Heewoo Jun, Tomer Kaftan, Łukasz Kaiser, Ali Kamali, Ingmar Kanitscheider, Nitish~Shirish Keskar, Tabarak Khan, Logan Kilpatrick, Jong~Wook Kim, Christina Kim, Yongjik Kim, Jan~Hendrik Kirchner, Jamie Kiros, Matt Knight, Daniel Kokotajlo, Łukasz Kondraciuk, Andrew Kondrich, Aris Konstantinidis, Kyle Kosic, Gretchen Krueger, Vishal Kuo, Michael Lampe, Ikai Lan, Teddy Lee, Jan Leike, Jade Leung, Daniel Levy, Chak~Ming Li, Rachel Lim, Molly Lin, Stephanie Lin, Mateusz Litwin, Theresa Lopez, Ryan
  Lowe, Patricia Lue, Anna Makanju, Kim Malfacini, Sam Manning, Todor Markov, Yaniv Markovski, Bianca Martin, Katie Mayer, Andrew Mayne, Bob McGrew, Scott~Mayer McKinney, Christine McLeavey, Paul McMillan, Jake McNeil, David Medina, Aalok Mehta, Jacob Menick, Luke Metz, Andrey Mishchenko, Pamela Mishkin, Vinnie Monaco, Evan Morikawa, Daniel Mossing, Tong Mu, Mira Murati, Oleg Murk, David Mély, Ashvin Nair, Reiichiro Nakano, Rajeev Nayak, Arvind Neelakantan, Richard Ngo, Hyeonwoo Noh, Long Ouyang, Cullen O'Keefe, Jakub Pachocki, Alex Paino, Joe Palermo, Ashley Pantuliano, Giambattista Parascandolo, Joel Parish, Emy Parparita, Alex Passos, Mikhail Pavlov, Andrew Peng, Adam Perelman, Filipe de~Avila Belbute~Peres, Michael Petrov, Henrique~Ponde de~Oliveira~Pinto, Michael, Pokorny, Michelle Pokrass, Vitchyr~H. Pong, Tolly Powell, Alethea Power, Boris Power, Elizabeth Proehl, Raul Puri, Alec Radford, Jack Rae, Aditya Ramesh, Cameron Raymond, Francis Real, Kendra Rimbach, Carl Ross, Bob Rotsted, Henri Roussez,
  Nick Ryder, Mario Saltarelli, Ted Sanders, Shibani Santurkar, Girish Sastry, Heather Schmidt, David Schnurr, John Schulman, Daniel Selsam, Kyla Sheppard, Toki Sherbakov, Jessica Shieh, Sarah Shoker, Pranav Shyam, Szymon Sidor, Eric Sigler, Maddie Simens, Jordan Sitkin, Katarina Slama, Ian Sohl, Benjamin Sokolowsky, Yang Song, Natalie Staudacher, Felipe~Petroski Such, Natalie Summers, Ilya Sutskever, Jie Tang, Nikolas Tezak, Madeleine~B. Thompson, Phil Tillet, Amin Tootoonchian, Elizabeth Tseng, Preston Tuggle, Nick Turley, Jerry Tworek, Juan Felipe~Cerón Uribe, Andrea Vallone, Arun Vijayvergiya, Chelsea Voss, Carroll Wainwright, Justin~Jay Wang, Alvin Wang, Ben Wang, Jonathan Ward, Jason Wei, CJ~Weinmann, Akila Welihinda, Peter Welinder, Jiayi Weng, Lilian Weng, Matt Wiethoff, Dave Willner, Clemens Winter, Samuel Wolrich, Hannah Wong, Lauren Workman, Sherwin Wu, Jeff Wu, Michael Wu, Kai Xiao, Tao Xu, Sarah Yoo, Kevin Yu, Qiming Yuan, Wojciech Zaremba, Rowan Zellers, Chong Zhang, Marvin Zhang, Shengjia
  Zhao, Tianhao Zheng, Juntang Zhuang, William Zhuk, and Barret Zoph. 2024.
\newblock \href {https://arxiv.org/abs/2303.08774} {Gpt-4 technical report}.
\newblock Technical report, OpenAI.

\bibitem[{Papineni et~al.(2002)Papineni, Roukos, Ward, and Zhu}]{papineni-etal-2002-bleu}
Kishore Papineni, Salim Roukos, Todd Ward, and Wei-Jing Zhu. 2002.
\newblock \href {https://doi.org/10.3115/1073083.1073135} {{B}leu: a method for automatic evaluation of machine translation}.
\newblock In \emph{Proceedings of the 40th Annual Meeting of the Association for Computational Linguistics}, pages 311--318, Philadelphia, Pennsylvania, USA. Association for Computational Linguistics.

\bibitem[{Perrella et~al.(2024)Perrella, Proietti, Scir{\`e}, Barba, and Navigli}]{perrella-etal-2024-guardians}
Stefano Perrella, Lorenzo Proietti, Alessandro Scir{\`e}, Edoardo Barba, and Roberto Navigli. 2024.
\newblock \href {https://doi.org/10.18653/v1/2024.acl-long.856} {Guardians of the machine translation meta-evaluation: Sentinel metrics fall in!}
\newblock In \emph{Proceedings of the 62nd Annual Meeting of the Association for Computational Linguistics (Volume 1: Long Papers)}, pages 16216--16244, Bangkok, Thailand. Association for Computational Linguistics.

\bibitem[{Perrella et~al.(2022)Perrella, Proietti, Scir{\`e}, Campolungo, and Navigli}]{perrella-etal-2022-matese}
Stefano Perrella, Lorenzo Proietti, Alessandro Scir{\`e}, Niccol{\`o} Campolungo, and Roberto Navigli. 2022.
\newblock \href {https://aclanthology.org/2022.wmt-1.51} {{M}a{TES}e: Machine translation evaluation as a sequence tagging problem}.
\newblock In \emph{Proceedings of the Seventh Conference on Machine Translation (WMT)}, pages 569--577, Abu Dhabi, United Arab Emirates (Hybrid). Association for Computational Linguistics.

\bibitem[{Peter et~al.(2023)Peter, Vilar, Deutsch, Finkelstein, Juraska, and Freitag}]{peter-etal-2023-theres}
Jan-Thorsten Peter, David Vilar, Daniel Deutsch, Mara Finkelstein, Juraj Juraska, and Markus Freitag. 2023.
\newblock \href {https://doi.org/10.18653/v1/2023.wmt-1.50} {There{'}s no data like better data: Using {QE} metrics for {MT} data filtering}.
\newblock In \emph{Proceedings of the Eighth Conference on Machine Translation}, pages 561--577, Singapore. Association for Computational Linguistics.

\bibitem[{Popovi{\'c}(2015)}]{popovic-2015-chrf}
Maja Popovi{\'c}. 2015.
\newblock \href {https://doi.org/10.18653/v1/W15-3049} {chr{F}: character n-gram {F}-score for automatic {MT} evaluation}.
\newblock In \emph{Proceedings of the Tenth Workshop on Statistical Machine Translation}, pages 392--395, Lisbon, Portugal. Association for Computational Linguistics.

\bibitem[{Popovi{\'c}(2017)}]{popovic-2017-chrf}
Maja Popovi{\'c}. 2017.
\newblock \href {https://doi.org/10.18653/v1/W17-4770} {chr{F}++: words helping character n-grams}.
\newblock In \emph{Proceedings of the Second Conference on Machine Translation}, pages 612--618, Copenhagen, Denmark. Association for Computational Linguistics.

\bibitem[{Pu et~al.(2021)Pu, Chung, Parikh, Gehrmann, and Sellam}]{pu-etal-2021-learning}
Amy Pu, Hyung~Won Chung, Ankur Parikh, Sebastian Gehrmann, and Thibault Sellam. 2021.
\newblock \href {https://doi.org/10.18653/v1/2021.emnlp-main.58} {Learning compact metrics for {MT}}.
\newblock In \emph{Proceedings of the 2021 Conference on Empirical Methods in Natural Language Processing}, pages 751--762, Online and Punta Cana, Dominican Republic. Association for Computational Linguistics.

\bibitem[{Ramos et~al.(2024)Ramos, Fernandes, Farinhas, and Martins}]{ramos-etal-2024-aligning}
Miguel Ramos, Patrick Fernandes, Ant{\'o}nio Farinhas, and Andre Martins. 2024.
\newblock \href {https://aclanthology.org/2024.eamt-1.22} {Aligning neural machine translation models: Human feedback in training and inference}.
\newblock In \emph{Proceedings of the 25th Annual Conference of the European Association for Machine Translation (Volume 1)}, pages 258--274, Sheffield, UK. European Association for Machine Translation (EAMT).

\bibitem[{Rei et~al.(2021)Rei, Farinha, Zerva, van Stigt, Stewart, Ramos, Glushkova, Martins, and Lavie}]{rei-etal-2021-references}
Ricardo Rei, Ana~C Farinha, Chrysoula Zerva, Daan van Stigt, Craig Stewart, Pedro Ramos, Taisiya Glushkova, Andr{\'e} F.~T. Martins, and Alon Lavie. 2021.
\newblock \href {https://aclanthology.org/2021.wmt-1.111} {Are references really needed? unbabel-{IST} 2021 submission for the metrics shared task}.
\newblock In \emph{Proceedings of the Sixth Conference on Machine Translation}, pages 1030--1040, Online. Association for Computational Linguistics.

\bibitem[{Rei et~al.(2023{\natexlab{a}})Rei, Guerreiro, Pombal, van Stigt, Treviso, Coheur, C.~de Souza, and Martins}]{rei-etal-2023-scaling}
Ricardo Rei, Nuno~M. Guerreiro, Jos{\~A}{\copyright} Pombal, Daan van Stigt, Marcos Treviso, Luisa Coheur, Jos{\'e}~G. C.~de Souza, and Andr{\'e} Martins. 2023{\natexlab{a}}.
\newblock \href {https://doi.org/10.18653/v1/2023.wmt-1.73} {Scaling up {C}omet{K}iwi: Unbabel-{IST} 2023 submission for the quality estimation shared task}.
\newblock In \emph{Proceedings of the Eighth Conference on Machine Translation}, pages 841--848, Singapore. Association for Computational Linguistics.

\bibitem[{Rei et~al.(2023{\natexlab{b}})Rei, Guerreiro, Treviso, Coheur, Lavie, and Martins}]{rei-etal-2023-inside}
Ricardo Rei, Nuno~M. Guerreiro, Marcos Treviso, Luisa Coheur, Alon Lavie, and Andr{\'e} Martins. 2023{\natexlab{b}}.
\newblock \href {https://doi.org/10.18653/v1/2023.acl-short.94} {The inside story: Towards better understanding of machine translation neural evaluation metrics}.
\newblock In \emph{Proceedings of the 61st Annual Meeting of the Association for Computational Linguistics (Volume 2: Short Papers)}, pages 1089--1105, Toronto, Canada. Association for Computational Linguistics.

\bibitem[{Rei et~al.(2020)Rei, Stewart, Farinha, and Lavie}]{rei-etal-2020-comet}
Ricardo Rei, Craig Stewart, Ana~C Farinha, and Alon Lavie. 2020.
\newblock \href {https://doi.org/10.18653/v1/2020.emnlp-main.213} {{COMET}: A neural framework for {MT} evaluation}.
\newblock In \emph{Proceedings of the 2020 Conference on Empirical Methods in Natural Language Processing (EMNLP)}, pages 2685--2702, Online. Association for Computational Linguistics.

\bibitem[{Rei et~al.(2022)Rei, Treviso, Guerreiro, Zerva, Farinha, Maroti, C.~de Souza, Glushkova, Alves, Coheur, Lavie, and Martins}]{rei-etal-2022-cometkiwi}
Ricardo Rei, Marcos Treviso, Nuno~M. Guerreiro, Chrysoula Zerva, Ana~C Farinha, Christine Maroti, Jos{\'e}~G. C.~de Souza, Taisiya Glushkova, Duarte Alves, Luisa Coheur, Alon Lavie, and Andr{\'e} F.~T. Martins. 2022.
\newblock \href {https://aclanthology.org/2022.wmt-1.60} {{C}omet{K}iwi: {IST}-unbabel 2022 submission for the quality estimation shared task}.
\newblock In \emph{Proceedings of the Seventh Conference on Machine Translation (WMT)}, pages 634--645, Abu Dhabi, United Arab Emirates (Hybrid). Association for Computational Linguistics.

\bibitem[{Sai~B et~al.(2023)Sai~B, Dixit, Nagarajan, Kunchukuttan, Kumar, Khapra, and Dabre}]{sai-b-etal-2023-indicmt}
Ananya Sai~B, Tanay Dixit, Vignesh Nagarajan, Anoop Kunchukuttan, Pratyush Kumar, Mitesh~M. Khapra, and Raj Dabre. 2023.
\newblock \href {https://doi.org/10.18653/v1/2023.acl-long.795} {{I}ndic{MT} eval: A dataset to meta-evaluate machine translation metrics for {I}ndian languages}.
\newblock In \emph{Proceedings of the 61st Annual Meeting of the Association for Computational Linguistics (Volume 1: Long Papers)}, pages 14210--14228, Toronto, Canada. Association for Computational Linguistics.

\bibitem[{Sellam et~al.(2020)Sellam, Das, and Parikh}]{sellam-etal-2020-bleurt}
Thibault Sellam, Dipanjan Das, and Ankur Parikh. 2020.
\newblock \href {https://doi.org/10.18653/v1/2020.acl-main.704} {{BLEURT}: Learning robust metrics for text generation}.
\newblock In \emph{Proceedings of the 58th Annual Meeting of the Association for Computational Linguistics}, pages 7881--7892, Online. Association for Computational Linguistics.

\bibitem[{Sennrich et~al.(2016)Sennrich, Haddow, and Birch}]{sennrich-etal-2016-neural}
Rico Sennrich, Barry Haddow, and Alexandra Birch. 2016.
\newblock \href {https://doi.org/10.18653/v1/P16-1162} {Neural machine translation of rare words with subword units}.
\newblock In \emph{Proceedings of the 54th Annual Meeting of the Association for Computational Linguistics (Volume 1: Long Papers)}, pages 1715--1725, Berlin, Germany. Association for Computational Linguistics.

\bibitem[{Stanojevi{\'c} et~al.(2015)Stanojevi{\'c}, Kamran, Koehn, and Bojar}]{stanojevic-etal-2015-results}
Milo{\v{s}} Stanojevi{\'c}, Amir Kamran, Philipp Koehn, and Ond{\v{r}}ej Bojar. 2015.
\newblock \href {https://doi.org/10.18653/v1/W15-3031} {Results of the {WMT}15 metrics shared task}.
\newblock In \emph{Proceedings of the Tenth Workshop on Statistical Machine Translation}, pages 256--273, Lisbon, Portugal. Association for Computational Linguistics.

\bibitem[{Team et~al.(2022)Team, Costa-jussà, Cross, Çelebi, Elbayad, Heafield, Heffernan, Kalbassi, Lam, Licht, Maillard, Sun, Wang, Wenzek, Youngblood, Akula, Barrault, Gonzalez, Hansanti, Hoffman, Jarrett, Sadagopan, Rowe, Spruit, Tran, Andrews, Ayan, Bhosale, Edunov, Fan, Gao, Goswami, Guzmán, Koehn, Mourachko, Ropers, Saleem, Schwenk, and Wang}]{nllbteam2022languageleftbehindscaling}
NLLB Team, Marta~R. Costa-jussà, James Cross, Onur Çelebi, Maha Elbayad, Kenneth Heafield, Kevin Heffernan, Elahe Kalbassi, Janice Lam, Daniel Licht, Jean Maillard, Anna Sun, Skyler Wang, Guillaume Wenzek, Al~Youngblood, Bapi Akula, Loic Barrault, Gabriel~Mejia Gonzalez, Prangthip Hansanti, John Hoffman, Semarley Jarrett, Kaushik~Ram Sadagopan, Dirk Rowe, Shannon Spruit, Chau Tran, Pierre Andrews, Necip~Fazil Ayan, Shruti Bhosale, Sergey Edunov, Angela Fan, Cynthia Gao, Vedanuj Goswami, Francisco Guzmán, Philipp Koehn, Alexandre Mourachko, Christophe Ropers, Safiyyah Saleem, Holger Schwenk, and Jeff Wang. 2022.
\newblock \href {https://arxiv.org/abs/2207.04672} {No language left behind: Scaling human-centered machine translation}.
\newblock \emph{arXiv preprint, arXiv:2207.04672}.

\bibitem[{Xu et~al.(2024)Xu, Sharaf, Chen, Tan, Shen, Durme, Murray, and Kim}]{xu2024contrastive}
Haoran Xu, Amr Sharaf, Yunmo Chen, Weiting Tan, Lingfeng Shen, Benjamin~Van Durme, Kenton Murray, and Young~Jin Kim. 2024.
\newblock \href {https://arxiv.org/abs/2401.08417} {Contrastive preference optimization: Pushing the boundaries of llm performance in machine translation}.
\newblock \emph{arXiv preprint, arXiv:2401.08417}.

\bibitem[{Xu et~al.(2023)Xu, Wang, Pan, Song, Freitag, Wang, and Li}]{xu-etal-2023-instructscore}
Wenda Xu, Danqing Wang, Liangming Pan, Zhenqiao Song, Markus Freitag, William Wang, and Lei Li. 2023.
\newblock \href {https://doi.org/10.18653/v1/2023.emnlp-main.365} {{INSTRUCTSCORE}: Towards explainable text generation evaluation with automatic feedback}.
\newblock In \emph{Proceedings of the 2023 Conference on Empirical Methods in Natural Language Processing}, pages 5967--5994, Singapore. Association for Computational Linguistics.

\bibitem[{Xue et~al.(2021)Xue, Constant, Roberts, Kale, Al-Rfou, Siddhant, Barua, and Raffel}]{xue-etal-2021-mt5}
Linting Xue, Noah Constant, Adam Roberts, Mihir Kale, Rami Al-Rfou, Aditya Siddhant, Aditya Barua, and Colin Raffel. 2021.
\newblock \href {https://doi.org/10.18653/v1/2021.naacl-main.41} {m{T}5: A massively multilingual pre-trained text-to-text transformer}.
\newblock In \emph{Proceedings of the 2021 Conference of the North American Chapter of the Association for Computational Linguistics: Human Language Technologies}, pages 483--498, Online. Association for Computational Linguistics.

\bibitem[{Zerva et~al.(2022)Zerva, Blain, Rei, Lertvittayakumjorn, C.~de Souza, Eger, Kanojia, Alves, Or{\u{a}}san, Fomicheva, Martins, and Specia}]{zerva-etal-2022-findings}
Chrysoula Zerva, Fr{\'e}d{\'e}ric Blain, Ricardo Rei, Piyawat Lertvittayakumjorn, Jos{\'e}~G. C.~de Souza, Steffen Eger, Diptesh Kanojia, Duarte Alves, Constantin Or{\u{a}}san, Marina Fomicheva, Andr{\'e} F.~T. Martins, and Lucia Specia. 2022.
\newblock \href {https://aclanthology.org/2022.wmt-1.3} {Findings of the {WMT} 2022 shared task on quality estimation}.
\newblock In \emph{Proceedings of the Seventh Conference on Machine Translation (WMT)}, pages 69--99, Abu Dhabi, United Arab Emirates (Hybrid). Association for Computational Linguistics.

\bibitem[{Zhang et~al.(2020)Zhang, Kishore, Wu, Weinberger, and Artzi}]{Zhang*2020BERTScore:}
Tianyi Zhang, Varsha Kishore, Felix Wu, Kilian~Q. Weinberger, and Yoav Artzi. 2020.
\newblock \href {https://openreview.net/forum?id=SkeHuCVFDr} {Bertscore: Evaluating text generation with bert}.
\newblock In \emph{Proceedings of the International Conference on Learning Representations}.

\end{thebibliography}

\appendix

\section{Implementation Details} \label{apx:implementation-details}
Most annotated datasets used for metric evaluation -- such as \wmtth and \wmttw -- contain a selection of source texts translated by multiple MT systems. As a result, each source text is paired with several automatic translations, along with one or more manually-curated references. In this respect, and to align the data filtering scenario to its real use case,\footnote{When used for data filtering, MT metrics filter parallel corpora that typically contain only one translation per source text.} we group the translations according to the MT system (or human annotator) that generated them, computing Precision and Recall on each group. In MT meta-evaluation, this strategy is called \textit{Group-by-System} or \textit{System Grouping}, by \citet{deutsch-etal-2023-ties} and  \citet{perrella-etal-2024-guardians}, respectively. Finally, we aggregate these statistics across systems obtaining average Precision and Recall measures, which are used to obtain the \fscore as in Equation~\ref{eq:fscore}. 

Instead, concerning the translation re-ranking scenario, translations are grouped according to their source text. This strategy is called \textit{Group-by-Item} or \textit{Segment Grouping} by \citet{deutsch-etal-2023-ties} and  \citet{perrella-etal-2024-guardians}, respectively. Consequently, the final Re-Ranking Precision is the average across the Precision values computed for each source text as in Equation~\ref{eq:accuracy}.

\paragraph{Selecting the Thresholds \texorpdfstring{$\bs{\tau}$}{tau}}
For each metric, we select the threshold $\tau$ that maximizes the \fscore, either on the test or development set. To find the optimal threshold for a metric, we i) collect all its assessments on the considered dataset, removing duplicates; ii) measure Precision, Recall, and \fscore corresponding to each candidate threshold value; and iii) select the threshold that yields the highest \fscore.

\section{Datasets Statistics} \label{apx:datasets-statistics}
Table~\ref{tab:datastats} presents the number of systems, segments, and annotations in \wmtth and \wmtthda.

Table~\ref{tab:mqm-statistics} presents the average and median MQM scores assigned to the translations in \wmtth and \wmttw.

\begin{table*}[t]
    \centering
    \begin{tabular}{lrrr|rrr|rrr}
    \toprule
        & \multicolumn{3}{c|}{\langpair{zh}{en}} & \multicolumn{3}{c|}{\langpair{en}{de}} & \multicolumn{3}{c}{\langpair{he}{en}} \\
        & \multicolumn{1}{c}{\textbf{Sys.}} & \multicolumn{1}{c}{\textbf{Seg.}} & \multicolumn{1}{c|}{\textbf{Total}} & \multicolumn{1}{c}{\textbf{Sys.}} & \multicolumn{1}{c}{\textbf{Seg.}} & \multicolumn{1}{c|}{\textbf{Total}} & \multicolumn{1}{c}{\textbf{Sys.}} & \multicolumn{1}{c}{\textbf{Seg.}} & \multicolumn{1}{c}{\textbf{Total}} \\
        \midrule
        \wmtth & $15$ & $1177$ & $17655$ & $12$ & $460$ & $5520$ & $13$ & $820$ & $10660$ \\
        \wmtthda & $15$ & $884$ & $13260$ & $12$ & $460$ & $5520$ & -- & -- & -- \\
    \bottomrule
    \end{tabular}
    \caption{Number of MT systems, source segments, and the total number of annotations in \wmtth and \wmtthda, excluding official WMT23 references employed by reference-based metrics. Concerning \wmtthda, we restricted the annotations to those available in \wmtth and discarded the rest.}
    \label{tab:datastats}
\end{table*}

\begin{table*}[]
    \centering
    \begin{tabular}{lrr|rr|rr}
    \toprule
    & \multicolumn{2}{c|}{\langpair{zh}{en}} & \multicolumn{2}{c|}{\langpair{en}{de}} & \multicolumn{2}{c}{\langpair{he}{en}} \\
    & \textbf{Avg.} & \textbf{Median} & \textbf{Avg.} & \textbf{Median} & \textbf{Avg.} & \textbf{Median} \\
    \midrule
    \wmtth  & $-4.21$ & $-2.00$ & $-7.47$ & $-3.00$ & $-2.35$ & $0.00$ \\
    \wmttw  & $-3.21$ & $-1.00$ & $-1.31$ & $0.00$ & -- & -- \\
    \bottomrule
    \end{tabular}
    \caption{Average and Median MQM scores of the translations in \wmtth and \wmttw.}
    \label{tab:mqm-statistics}
\end{table*}

\section{The Metrics} \label{apx:metrics}
We consider the following metrics:
\begin{itemize}
    \item \comet, \cometqe, and \cometmqmqe, \cite{rei-etal-2020-comet, rei-etal-2021-references} are a reference-based and two reference-free regression-based metrics, respectively, built upon the XLM-RoBERTa large architecture \cite{conneau-etal-2020-unsupervised}, and trained using datasets containing human annotations in the form of Direct Assessments (DA) \cite{graham-etal-2013-continuous}. Specifically, \comet was trained on the datasets released at WMT between $2017$ and $2020$ \cite{bojar-etal-2017-results, ma-etal-2018-results, ma-etal-2019-results, mathur-etal-2020-results}, while \cometqe and \cometmqmqe also include the DA-based datasets released in $2015$ and $2016$ \cite{stanojevic-etal-2015-results, bojar-etal-2016-results}. \cometmqmqe was further fine-tuned on a split of the MQM-based corpus released by \citet{freitag-etal-2021-experts}.\footnote{\label{fn:comet}\url{https://github.com/Unbabel/COMET}. We used the version 2.2.1 of the COMET framework.} 

    \item \bleurt \cite{sellam-etal-2020-bleurt, pu-etal-2021-learning} is a regression-based metric built upon the RemBERT pre-trained language model \cite{chung2021rethinking}. RemBERT was fine-tuned on DA-based human assessments from $2015$ to $2019$, along with synthetic data.\footnote{\url{https://github.com/google-research/bleurt}.}

    \item \bertscore \cite{Zhang*2020BERTScore:} leverages pre-trained encoders to extract the contextualized embeddings of the tokens of a translation and its reference. Then, it computes the cosine similarity between each pair of embeddings, greedily matching the most similar ones.\footnote{\url{https://github.com/Tiiiger/bert_score}.}

    \item \metricx, \metricxqe, \metricxxl and \metricxxlqe \cite{juraska-etal-2023-metricx} are regression-based metrics built upon the mT5-XXL (the first two) and mT5-XL (the last two) models \cite{xue-etal-2021-mt5}. These metrics are trained using DA-based human judgments released at WMT between $2015$ and $2020$ \cite{stanojevic-etal-2015-results, bojar-etal-2016-results}, and further fine-tuned on a combination of MQM-based human judgments and synthetic data \cite{freitag-etal-2021-experts, freitag-etal-2021-results}.\footnote{\url{https://github.com/google-research/metricx}.} 

    \item \cometkiwi and \cometkiwixl \cite{rei-etal-2022-cometkiwi, rei-etal-2023-scaling} are reference-free regression-based metrics, built upon InfoXLM \cite{chi-etal-2021-infoxlm} and XLM-RoBERTa XL \cite{goyal-etal-2021-larger}, respectively. These metrics are trained using DA-based human judgments released at WMT from $2017$ to $2020$, as well as DA from the MLQE-PE corpus \cite{fomicheva-etal-2022-mlqe}. \cometkiwixl's training data also include the DA for Indian languages released by \citet{blain-etal-2023-findings}. \footnote{See footnote \ref{fn:comet}.}

     \item \xcometxl \cite{guerreiro2023xcomet} is a regression-based metric built upon the XLM-RoBERTa XL architecture, trained on the concatenation of DA-based human judgments released at WMT from $2017$ and $2020$ and the MLQE-PE dataset, and further fine-tuned using MQM-based annotations coming from the following datasets: i) WMT data from $2020$ to $2022$, ii) IndicMT \cite{sai-b-etal-2023-indicmt}, and iii) DEMETR \cite{karpinska-etal-2022-demetr}. Given a candidate translation, \xcometxl jointly identifies its error spans and assigns it a scalar quality score. \xcometens and \xcometensqe are ensembles between one XL and two XXL \xcomet checkpoints that result from different training stages.\footnote{See footnote \ref{fn:comet}.}

    \item \matese and \mateseqe \cite{perrella-etal-2022-matese} are a reference-based and a reference-free metric, respectively, built upon InfoXLM and DeBERTaV3 \cite{he2023debertav3}. \matese metrics annotate the spans of translations that contain an error, specifying the error severity.\footnote{\url{https://github.com/SapienzaNLP/MaTESe}.} 

    \item \gemba \cite{kocmi-federmann-2023-gemba} is an LLM-based metric that leverages GPT-4 to return quality assessments in the form of MQM annotations.\footnote{\url{https://github.com/MicrosoftTranslator/GEMBA}.}

    \item \mbrmetricxqe \cite{naskar-etal-2023-quality} is based on the MBR decoding strategy. Given a translation, it uses an MT system to generate pseudo-references and a reference-based MT metric (\metricx) as the MBR utility function.

    \item \bleu \cite{papineni-etal-2002-bleu} is a precision-oriented metric that computes the number of overlapping n-grams between a translation and its reference.\footnote{\label{fn:bleu}\url{https://github.com/mjpost/sacrebleu}.}

    \item \chrf \cite{popovic-2015-chrf} compares a translation and its reference based on the number of overlapping character n-grams.\footnote{See footnote \ref{fn:bleu}.}

    \item \fbleu \cite{goyal-etal-2022-flores, nllbteam2022languageleftbehindscaling} computes BLEU scores using subword tokenization done by the standardized FLORES-200 Sentencepiece models.\footnote{See footnote \ref{fn:bleu}.}

    \item \ebleu \cite{elnokrashy-kocmi-2023-ebleu} matches the n-grams of semantically similar words between a candidate translation and a reference using non-contextual word embeddings.\footnote{\url{https://github.com/munael/ebleu-mt-metrics-wmt23}.}

    \item \tokengramf \cite{dreano-etal-2023-tokengram} is derived from chrF++ \cite{popovic-2017-chrf} by replacing word-based n-grams with token-based n-grams, as obtained from popular tokenization algorithms such as BPE \cite{sennrich-etal-2016-neural} or Unigram \cite{kudo-2018-subword}.\footnote{\url{https://github.com/SorenDreano/tokengram_F}.}
   
\end{itemize}
In addition, we include three sentinel metrics, i.e., metrics designed explicitly to detect issues with the meta-evaluation \cite{perrella-etal-2024-guardians}:
\begin{itemize}
    \item \candonly assesses the quality of a translation without taking its source or reference as input.

    \item \srconly predicts the quality of a translation based solely on its source, without taking the translation itself as input.

    \item \refonly predicts the quality of a translation based solely on its reference, without taking the translation itself as input.
\end{itemize}
Trained with incomplete information, these metrics are not supposed to rank high in a fair meta-evaluation setup. Sentinel metrics are regression-based and were trained using WMT data. In particular, they were trained using DA annotations from $2017$ to $2020$ and further fine-tuned with MQM scores from $2020$ to $2022$.

\section{Additional Results} \label{apx:performance-results}

In this section, we report all our results considering all the language directions available in \wmtth, i.e., \langpair{zh}{en}, \langpair{en}{de}, and \langpair{he}{en}, and including all MT metrics mentioned in Appendix~\ref{apx:metrics}.

Tables~\ref{tab:performance-results-zhen-apx}, \ref{tab:performance-results-ende-apx}, and \ref{tab:performance-results-heen-apx} show the performance of MT metrics in the data filtering scenario when $\tau$ is selected as the one that maximizes \fscore on the test set, i.e., \wmtth. The last two columns contain the performance of MT metrics in the translation re-ranking scenario. 
Instead, Tables~\ref{tab:performance-results-zhen-dev-apx} and \ref{tab:performance-results-ende-dev-apx} show the performance of MT metrics in the data filtering scenario when $\tau$ is selected as the one that maximizes the \fscore on the development set, i.e., \wmttw, and the performance is measured on \wmtth.

\paragraph{The performance of lexical-based metrics}
All lexical-based metrics fail, partially or completely, at tackling the data filtering task. In most cases, their optimal threshold is $0.0$\footnote{Note that these metrics' score range is either $[0,1]$ or $[0,100]$.}, indicating that they lack the sensitivity required to separate \good from \bad and \perfect from \other translations, and therefore resort to maximizing recall. Instead, lexical-based metrics achieve a decent performance in the translation re-ranking scenario. Nonetheless, they still perform worse than most neural-based metrics.

\paragraph{The performance of sentinel metrics}
As illustrated in Appendix~\ref{apx:implementation-details}, we use the \textit{System Grouping} strategy to align the data filtering scenario to its real use case. Specifically, we compute Precision and Recall on the translations of each MT system independently and then compute final statistics by averaging them across MT systems. As demonstrated by \citet{perrella-etal-2024-guardians}, this setting is particularly susceptible to spurious correlations in the evaluation data, which favor trained metrics over the rest. As a consequence, the performance of sentinel metrics is not as low as it would be in a fair evaluation scenario. However, we highlight that this scenario was intentionally designed to adhere closely to the data filtering use case, and adopting a different grouping strategy could reduce its effectiveness as a proxy for this task. Therefore, we argue that careful attention should be given to selecting source texts in evaluation datasets, with the goal of minimizing the impact of spurious correlations and ultimately ensuring that sentinel metrics rank at the bottom of the metric rankings. Nonetheless, despite sentinel metrics performing better than they ideally ought to, they still do not surpass most state-of-the-art metrics, differently from the results obtained by \citet{perrella-etal-2024-guardians}. Similarly, \gemba performs decently in many of our settings, whereas \citet{perrella-etal-2024-guardians} report it ranking lower than sentinel metrics when using \textit{System Grouping} (specifically, in two out of three translation directions, namely \langpair{zh}{en} and \langpair{en}{de}).\footnote{Since \gemba was not fine-tuned using human assessments, it should not be able to leverage spurious correlations in metrics' training data to conduct the evaluation. \citet{perrella-etal-2024-guardians} report \gemba ranking lower than sentinel metrics, suggesting that the evaluation might unfairly favor metrics that have learned spurious correlations during training.} Given these observations, we believe that the binary classification setup lessens the impact of spurious correlations, as compared to the correlation with human judgment.

Instead, and as expected, sentinel metrics rank at the bottom in translation re-ranking. Indeed, the translation re-ranking scenario involves selecting the best among the translations of the same source text, i.e., using the \textit{Segment Grouping} strategy, which, as shown by \citet{perrella-etal-2024-guardians}, counters the impact of spurious correlations in the evaluation dataset.

\begin{table*}[t]
\centering
\resizebox{\linewidth}{!}{
    \begin{NiceTabular}{ll|rrrr|rrrr|rr}[cell-space-limits=3pt]
    \toprule
     &  & \multicolumn{4}{c|}{\textbf{\good vs \bad}} & \multicolumn{4}{c|}{\textbf{\perfect vs \other}} & \multicolumn{2}{c}{\textbf{Re-ranking}} \\
    & \textbf{Metric} & \multicolumn{1}{c}{$\boldsymbol{\tau}$} & \multicolumn{1}{c}{\textbf{P}} & \multicolumn{1}{c}{\textbf{R}} & \multicolumn{1}{c|}{$\boldsymbol{F}$} & \multicolumn{1}{c}{$\boldsymbol{\tau}$} & \multicolumn{1}{c}{\textbf{P}} & \multicolumn{1}{c}{\textbf{R}} & \multicolumn{1}{c|}{$\boldsymbol{F}$} & \multicolumn{1}{c}{\textbf{RRP}} & \multicolumn{1}{c}{\textbf{Avg.}} \\
    \cmidrule(lr){2-12}
    \multirow{ 8 }{*}{\rotatebox{90}{\shortstack{\textsc{reference} \\ \textsc{based}}}}
    & \cellcolor{highlight}\xcometens & \cellcolor{highlight}$0.83$ & \cellcolor{highlight}$79.91$ & \cellcolor{highlight}$84.42$ & \cellcolor{highlight}$81.36$ & \cellcolor{highlight}$0.91$ & \cellcolor{highlight}$68.25$ & \cellcolor{highlight}$68.93$ & \cellcolor{highlight}$68.47$ & \cellcolor{highlight}$43.17$ & \cellcolor{highlight}$-2.38$ \\
    & \xcometxl & $0.80$ & $78.33$ & $83.63$ & $80.02$ & $0.92$ & $67.55$ & $67.46$ & $67.52$ & $37.49$ & $-2.75$ \\
    & \cellcolor{highlight}\metricx & \cellcolor{highlight}$-4.79$ & \cellcolor{highlight}$77.43$ & \cellcolor{highlight}$86.23$ & \cellcolor{highlight}$80.15$ & \cellcolor{highlight}$-2.25$ & \cellcolor{highlight}$63.99$ & \cellcolor{highlight}$73.20$ & \cellcolor{highlight}$66.79$ & \cellcolor{highlight}$39.63$ & \cellcolor{highlight}$-2.72$ \\
    & \metricxxl & $-3.52$ & $77.80$ & $84.46$ & $79.90$ & $-1.74$ & $65.60$ & $72.54$ & $67.76$ & $39.52$ & $-2.71$ \\
    & \matese & $-4.00$ & $76.53$ & $78.10$ & $77.05$ & $-1.00$ & $55.75$ & $79.88$ & $61.99$ & $33.07$ & $-3.18$ \\
    & \comet & $0.76$ & $74.56$ & $78.76$ & $75.91$ & $0.82$ & $61.25$ & $64.38$ & $62.26$ & $34.25$ & $-3.06$ \\
    & \bleurt & $0.60$ & $72.76$ & $82.76$ & $75.81$ & $0.67$ & $55.88$ & $69.21$ & $59.71$ & $33.35$ & $-3.07$ \\
    & \bertscore & $0.84$ & $64.33$ & $99.47$ & $72.91$ & $0.92$ & $48.20$ & $69.15$ & $53.62$ & $32.29$ & $-3.20$ \\
    \cmidrule(lr){2-12}
    \multirow{ 10 }{*}{\rotatebox{90}{\shortstack{\textsc{reference} \\ \textsc{free}}}}
    & \cellcolor{highlight}\xcometensqe & \cellcolor{highlight}$0.83$ & \cellcolor{highlight}$80.40$ & \cellcolor{highlight}$83.47$ & \cellcolor{highlight}$81.40$ & \cellcolor{highlight}$0.92$ & \cellcolor{highlight}$70.00$ & \cellcolor{highlight}$63.60$ & \cellcolor{highlight}$67.73$ & \cellcolor{highlight}$41.40$ & \cellcolor{highlight}$-2.47$ \\
    & \cellcolor{highlight}\mbrmetricxqe & \cellcolor{highlight}$0.73$ & \cellcolor{highlight}$79.00$ & \cellcolor{highlight}$82.81$ & \cellcolor{highlight}$80.23$ & \cellcolor{highlight}$0.80$ & \cellcolor{highlight}$67.02$ & \cellcolor{highlight}$65.91$ & \cellcolor{highlight}$66.64$ & \cellcolor{highlight}$38.47$ & \cellcolor{highlight}$-2.40$ \\
    & \cellcolor{highlight}\metricxqe & \cellcolor{highlight}$-3.90$ & \cellcolor{highlight}$76.73$ & \cellcolor{highlight}$87.70$ & \cellcolor{highlight}$80.07$ & \cellcolor{highlight}$-1.31$ & \cellcolor{highlight}$67.76$ & \cellcolor{highlight}$67.85$ & \cellcolor{highlight}$67.79$ & \cellcolor{highlight}$37.55$ & \cellcolor{highlight}$-2.59$ \\
    & \metricxxlqe & $-3.57$ & $77.91$ & $83.36$ & $79.64$ & $-1.64$ & $67.15$ & $70.08$ & $68.10$ & $36.09$ & $-2.83$ \\
    & \cellcolor{highlight}\gemba & \cellcolor{highlight}$-5.00$ & \cellcolor{highlight}$82.41$ & \cellcolor{highlight}$79.99$ & \cellcolor{highlight}$81.59$ & \cellcolor{highlight}$-1.00$ & \cellcolor{highlight}$64.12$ & \cellcolor{highlight}$74.12$ & \cellcolor{highlight}$67.14$ & \cellcolor{highlight}$42.58$ & \cellcolor{highlight}$-2.30$ \\
    & \mateseqe & $-4.00$ & $73.72$ & $85.64$ & $77.30$ & $0.00$ & $55.43$ & $75.05$ & $60.72$ & $30.34$ & $-3.59$ \\
    & \cometqe & $-0.01$ & $75.35$ & $82.53$ & $77.60$ & $0.05$ & $59.64$ & $68.59$ & $62.35$ & $37.35$ & $-2.66$ \\
    & \cometmqmqe & $0.08$ & $75.40$ & $86.33$ & $78.72$ & $0.10$ & $61.63$ & $73.84$ & $65.22$ & $33.52$ & $-3.59$ \\
    & \cometkiwi & $0.76$ & $78.62$ & $80.90$ & $79.37$ & $0.80$ & $64.79$ & $66.52$ & $65.35$ & $39.28$ & $-2.61$ \\
    & \cometkiwixl & $0.64$ & $78.04$ & $79.81$ & $78.62$ & $0.71$ & $64.73$ & $65.51$ & $64.99$ & $38.78$ & $-2.60$ \\
    \cmidrule(lr){2-12}
    \multirow{ 5 }{*}{\rotatebox{90}{\shortstack{\textsc{lexical} \\ \textsc{based}}}}
    & \bleu & $0.00$ & $64.06$ & $100.00$ & $72.78$ & $0.00$ & $42.13$ & $100.00$ & $52.20$ & $30.09$ & $-3.50$ \\
    & \chrf & $0.00$ & $64.06$ & $100.00$ & $72.78$ & $0.00$ & $42.13$ & $100.00$ & $52.20$ & $31.51$ & $-3.39$ \\
    & \ebleu & $0.02$ & $64.11$ & $99.82$ & $72.79$ & $0.03$ & $42.20$ & $99.87$ & $52.26$ & $30.16$ & $-3.49$ \\
    & \fbleu & $0.00$ & $64.06$ & $100.00$ & $72.78$ & $0.00$ & $42.13$ & $100.00$ & $52.20$ & $30.80$ & $-3.46$ \\
    & \tokengramf & $0.00$ & $64.06$ & $100.00$ & $72.78$ & $0.00$ & $42.13$ & $100.00$ & $52.20$ & $30.55$ & $-3.44$ \\
    \cmidrule(lr){2-12}
    \multirow{ 3 }{*}{\rotatebox{90}{\shortstack{\textsc{sentinel} \\ \textsc{metrics}}}}
    & \srconly & $-0.14$ & $75.64$ & $83.31$ & $78.03$ & $0.23$ & $63.00$ & $71.90$ & $65.71$ & $25.77$ & $-4.21$ \\
    & \refonly & $-0.55$ & $71.74$ & $91.25$ & $77.24$ & $0.08$ & $59.11$ & $73.33$ & $63.19$ & $25.77$ & $-4.21$ \\
    & \candonly & $-0.14$ & $75.43$ & $86.92$ & $78.91$ & $0.22$ & $63.16$ & $71.84$ & $65.81$ & $29.38$ & $-3.83$ \\
    \cmidrule(lr){2-12}
    & \randombaseline & $-5.00$ & $64.06$ & $100.00$ & $72.78$ & $-4.00$ & $42.14$ & $99.99$ & $52.21$ & $29.04$ & $-3.74$ \\
    & \dasqm & $63.50$ & $67.83$ & $95.95$ & $75.18$ & $74.67$ & $48.30$ & $82.61$ & $56.06$ & $32.99$ & $-3.22$ \\
    \bottomrule
    \end{NiceTabular}
}

\caption{Metrics' Precision, Recall, and \fscore in binary classification, distinguishing \good from \bad, and \perfect from \other translations. $\tau$ is selected to maximize the \fscore \textbf{on the test set}. In the last two columns, we report metrics' Precision in translation re-ranking and the average MQM score of the selected translations. The test set is \wmtth and the translation direction is \langpair{zh}{en}. The metrics highlighted in grey are not openly available.}
\label{tab:performance-results-zhen-apx}
\end{table*}

\begin{table*}[t]
\centering
\resizebox{\linewidth}{!}{
    \begin{NiceTabular}{ll|rrrr|rrrr|rr}[cell-space-limits=3pt]
    \toprule
     &  & \multicolumn{4}{c|}{\textbf{\good vs \bad}} & \multicolumn{4}{c|}{\textbf{\perfect vs \other}} & \multicolumn{2}{c}{\textbf{Re-ranking}} \\
    & \textbf{Metric} & \multicolumn{1}{c}{$\boldsymbol{\tau}$} & \multicolumn{1}{c}{\textbf{P}} & \multicolumn{1}{c}{\textbf{R}} & \multicolumn{1}{c|}{$\boldsymbol{F}$} & \multicolumn{1}{c}{$\boldsymbol{\tau}$} & \multicolumn{1}{c}{\textbf{P}} & \multicolumn{1}{c}{\textbf{R}} & \multicolumn{1}{c|}{$\boldsymbol{F}$} & \multicolumn{1}{c}{\textbf{RRP}} & \multicolumn{1}{c}{\textbf{Avg.}} \\
    \cmidrule(lr){2-12}
    \multirow{ 8 }{*}{\rotatebox{90}{\shortstack{\textsc{reference} \\ \textsc{based}}}}
    & \cellcolor{highlight}\xcometens & \cellcolor{highlight}$0.90$ & \cellcolor{highlight}$80.55$ & \cellcolor{highlight}$70.52$ & \cellcolor{highlight}$76.90$ & \cellcolor{highlight}$0.92$ & \cellcolor{highlight}$75.90$ & \cellcolor{highlight}$67.86$ & \cellcolor{highlight}$73.02$ & \cellcolor{highlight}$48.79$ & \cellcolor{highlight}$-3.58$ \\
    & \xcometxl & $0.90$ & $78.49$ & $71.17$ & $75.89$ & $0.94$ & $78.17$ & $65.44$ & $73.41$ & $47.31$ & $-3.91$ \\
    & \cellcolor{highlight}\metricx & \cellcolor{highlight}$-1.71$ & \cellcolor{highlight}$79.56$ & \cellcolor{highlight}$67.61$ & \cellcolor{highlight}$75.14$ & \cellcolor{highlight}$-1.18$ & \cellcolor{highlight}$75.21$ & \cellcolor{highlight}$68.18$ & \cellcolor{highlight}$72.71$ & \cellcolor{highlight}$52.61$ & \cellcolor{highlight}$-3.47$ \\
    & \metricxxl & $-1.55$ & $77.62$ & $73.98$ & $76.37$ & $-1.07$ & $72.59$ & $70.88$ & $72.01$ & $47.81$ & $-3.74$ \\
    & \matese & $-1.00$ & $71.87$ & $72.92$ & $72.21$ & $0.00$ & $67.42$ & $60.67$ & $65.01$ & $43.18$ & $-4.56$ \\
    & \comet & $0.83$ & $72.56$ & $70.81$ & $71.97$ & $0.88$ & $81.10$ & $55.66$ & $70.38$ & $48.26$ & $-3.50$ \\
    & \bleurt & $0.70$ & $76.49$ & $69.54$ & $74.03$ & $0.74$ & $72.66$ & $63.26$ & $69.23$ & $48.27$ & $-3.64$ \\
    & \bertscore & $0.85$ & $57.66$ & $81.22$ & $63.83$ & $0.92$ & $68.12$ & $40.42$ & $55.45$ & $43.11$ & $-4.55$ \\
    \cmidrule(lr){2-12}
    \multirow{ 10 }{*}{\rotatebox{90}{\shortstack{\textsc{reference} \\ \textsc{free}}}}
    & \cellcolor{highlight}\xcometensqe & \cellcolor{highlight}$0.87$ & \cellcolor{highlight}$79.99$ & \cellcolor{highlight}$70.39$ & \cellcolor{highlight}$76.51$ & \cellcolor{highlight}$0.91$ & \cellcolor{highlight}$75.58$ & \cellcolor{highlight}$66.10$ & \cellcolor{highlight}$72.13$ & \cellcolor{highlight}$46.70$ & \cellcolor{highlight}$-3.90$ \\
    & \cellcolor{highlight}\mbrmetricxqe & \cellcolor{highlight}$0.76$ & \cellcolor{highlight}$78.82$ & \cellcolor{highlight}$70.50$ & \cellcolor{highlight}$75.84$ & \cellcolor{highlight}$0.80$ & \cellcolor{highlight}$74.25$ & \cellcolor{highlight}$66.75$ & \cellcolor{highlight}$71.57$ & \cellcolor{highlight}$48.81$ & \cellcolor{highlight}$-3.78$ \\
    & \cellcolor{highlight}\metricxqe & \cellcolor{highlight}$-2.07$ & \cellcolor{highlight}$75.65$ & \cellcolor{highlight}$75.84$ & \cellcolor{highlight}$75.71$ & \cellcolor{highlight}$-1.09$ & \cellcolor{highlight}$76.81$ & \cellcolor{highlight}$64.88$ & \cellcolor{highlight}$72.37$ & \cellcolor{highlight}$48.04$ & \cellcolor{highlight}$-3.58$ \\
    & \metricxxlqe & $-1.99$ & $75.86$ & $73.18$ & $74.94$ & $-1.27$ & $74.08$ & $68.44$ & $72.10$ & $45.57$ & $-3.96$ \\
    & \cellcolor{highlight}\gemba & \cellcolor{highlight}$-1.00$ & \cellcolor{highlight}$79.69$ & \cellcolor{highlight}$66.77$ & \cellcolor{highlight}$74.86$ & \cellcolor{highlight}$0.00$ & \cellcolor{highlight}$75.07$ & \cellcolor{highlight}$62.04$ & \cellcolor{highlight}$70.16$ & \cellcolor{highlight}$42.52$ & \cellcolor{highlight}$-4.04$ \\
    & \mateseqe & $-2.00$ & $67.48$ & $82.89$ & $71.93$ & $0.00$ & $68.16$ & $63.48$ & $66.52$ & $41.03$ & $-5.14$ \\
    & \cometqe & $0.04$ & $68.75$ & $73.61$ & $70.30$ & $0.07$ & $65.26$ & $62.99$ & $64.49$ & $45.71$ & $-3.84$ \\
    & \cometmqmqe & $0.08$ & $74.18$ & $73.98$ & $74.12$ & $0.09$ & $73.50$ & $65.64$ & $70.68$ & $41.25$ & $-4.82$ \\
    & \cometkiwi & $0.82$ & $75.86$ & $68.51$ & $73.24$ & $0.82$ & $64.08$ & $71.37$ & $66.34$ & $41.75$ & $-4.32$ \\
    & \cometkiwixl & $0.69$ & $73.52$ & $70.71$ & $72.56$ & $0.73$ & $67.44$ & $64.13$ & $66.30$ & $43.67$ & $-4.45$ \\
    \cmidrule(lr){2-12}
    \multirow{ 5 }{*}{\rotatebox{90}{\shortstack{\textsc{lexical} \\ \textsc{based}}}}
    & \bleu & $3.29$ & $52.19$ & $99.25$ & $61.99$ & $0.00$ & $39.95$ & $100.00$ & $49.94$ & $42.95$ & $-4.28$ \\
    & \chrf & $28.67$ & $52.59$ & $99.47$ & $62.39$ & $73.11$ & $62.32$ & $39.61$ & $52.32$ & $41.43$ & $-4.48$ \\
    & \ebleu & $0.14$ & $52.12$ & $99.63$ & $61.97$ & $0.75$ & $62.15$ & $37.38$ & $50.90$ & $40.86$ & $-4.76$ \\
    & \fbleu & $7.02$ & $52.56$ & $98.56$ & $62.25$ & $48.17$ & $53.42$ & $51.43$ & $52.74$ & $43.69$ & $-4.21$ \\
    & \tokengramf & $0.29$ & $52.60$ & $99.18$ & $62.36$ & $0.69$ & $54.63$ & $47.20$ & $51.91$ & $42.63$ & $-4.43$ \\
    \cmidrule(lr){2-12}
    \multirow{ 3 }{*}{\rotatebox{90}{\shortstack{\textsc{sentinel} \\ \textsc{metrics}}}}
    & \srconly & $0.22$ & $75.96$ & $69.88$ & $73.82$ & $0.37$ & $73.80$ & $62.93$ & $69.78$ & $30.98$ & $-7.47$ \\
    & \refonly & $0.17$ & $73.12$ & $68.86$ & $71.65$ & $0.27$ & $68.76$ & $66.90$ & $68.13$ & $30.98$ & $-7.47$ \\
    & \candonly & $0.20$ & $75.73$ & $68.80$ & $73.27$ & $0.28$ & $68.70$ & $69.00$ & $68.80$ & $43.93$ & $-4.72$ \\
    \cmidrule(lr){2-12}
    & \randombaseline & $-4.00$ & $52.07$ & $99.94$ & $61.96$ & $-4.00$ & $39.96$ & $99.92$ & $49.95$ & $40.56$ & $-5.36$ \\
    & \dasqm & $77.33$ & $59.60$ & $85.39$ & $66.27$ & $82.67$ & $48.90$ & $77.21$ & $55.71$ & $37.11$ & $-5.01$ \\
    \bottomrule
    \end{NiceTabular}
}

\caption{Metrics' Precision, Recall, and \fscore in binary classification, distinguishing \good from \bad, and \perfect from \other translations. $\tau$ is selected to maximize the \fscore \textbf{on the test set}. In the last two columns, we report metrics' Precision in translation re-ranking and the average MQM score of the selected translations. The test set is \wmtth and the translation direction is \langpair{en}{de}. The metrics highlighted in grey are not openly available.}
\label{tab:performance-results-ende-apx}
\end{table*}

\begin{table*}[t]
\centering
\resizebox{\linewidth}{!}{
    \begin{NiceTabular}{ll|rrrr|rrrr|rr}[cell-space-limits=3pt]
    \toprule
     &  & \multicolumn{4}{c|}{\textbf{\good vs \bad}} & \multicolumn{4}{c|}{\textbf{\perfect vs \other}} & \multicolumn{2}{c}{\textbf{Re-ranking}} \\
    & \textbf{Metric} & \multicolumn{1}{c}{$\boldsymbol{\tau}$} & \multicolumn{1}{c}{\textbf{P}} & \multicolumn{1}{c}{\textbf{R}} & \multicolumn{1}{c|}{$\boldsymbol{F}$} & \multicolumn{1}{c}{$\boldsymbol{\tau}$} & \multicolumn{1}{c}{\textbf{P}} & \multicolumn{1}{c}{\textbf{R}} & \multicolumn{1}{c|}{$\boldsymbol{F}$} & \multicolumn{1}{c}{\textbf{RRP}} & \multicolumn{1}{c}{\textbf{Avg.}} \\
    \cmidrule(lr){2-12}
    \multirow{ 8 }{*}{\rotatebox{90}{\shortstack{\textsc{reference} \\ \textsc{based}}}}
    & \cellcolor{highlight}\xcometens & \cellcolor{highlight}$0.84$ & \cellcolor{highlight}$83.28$ & \cellcolor{highlight}$85.81$ & \cellcolor{highlight}$84.10$ & \cellcolor{highlight}$0.87$ & \cellcolor{highlight}$83.42$ & \cellcolor{highlight}$80.69$ & \cellcolor{highlight}$82.49$ & \cellcolor{highlight}$69.21$ & \cellcolor{highlight}$-0.99$ \\
    & \xcometxl & $0.81$ & $82.00$ & $87.14$ & $83.64$ & $0.85$ & $81.24$ & $82.82$ & $81.76$ & $68.31$ & $-0.98$ \\
    & \cellcolor{highlight}\metricx & \cellcolor{highlight}$-4.26$ & \cellcolor{highlight}$82.09$ & \cellcolor{highlight}$86.51$ & \cellcolor{highlight}$83.51$ & \cellcolor{highlight}$-3.34$ & \cellcolor{highlight}$81.78$ & \cellcolor{highlight}$81.96$ & \cellcolor{highlight}$81.84$ & \cellcolor{highlight}$67.20$ & \cellcolor{highlight}$-1.11$ \\
    & \metricxxl & $-3.44$ & $81.94$ & $87.23$ & $83.63$ & $-3.39$ & $79.11$ & $88.20$ & $81.92$ & $67.17$ & $-1.07$ \\
    & \matese & $-4.00$ & $84.49$ & $76.57$ & $81.67$ & $-4.00$ & $81.86$ & $77.99$ & $80.53$ & $61.99$ & $-1.51$ \\
    & \comet & $0.77$ & $78.96$ & $87.93$ & $81.74$ & $0.81$ & $78.95$ & $81.31$ & $79.72$ & $70.01$ & $-1.03$ \\
    & \bleurt & $0.67$ & $80.85$ & $83.39$ & $81.68$ & $0.67$ & $77.48$ & $84.02$ & $79.55$ & $68.33$ & $-1.04$ \\
    & \bertscore & $0.92$ & $76.33$ & $88.14$ & $79.90$ & $0.93$ & $73.89$ & $85.97$ & $77.52$ & $69.88$ & $-0.88$ \\
    \cmidrule(lr){2-12}
    \multirow{ 10 }{*}{\rotatebox{90}{\shortstack{\textsc{reference} \\ \textsc{free}}}}
    & \cellcolor{highlight}\xcometensqe & \cellcolor{highlight}$0.82$ & \cellcolor{highlight}$80.92$ & \cellcolor{highlight}$86.90$ & \cellcolor{highlight}$82.82$ & \cellcolor{highlight}$0.85$ & \cellcolor{highlight}$80.96$ & \cellcolor{highlight}$80.60$ & \cellcolor{highlight}$80.84$ & \cellcolor{highlight}$66.22$ & \cellcolor{highlight}$-1.40$ \\
    & \cellcolor{highlight}\mbrmetricxqe & \cellcolor{highlight}$0.74$ & \cellcolor{highlight}$81.57$ & \cellcolor{highlight}$86.17$ & \cellcolor{highlight}$83.05$ & \cellcolor{highlight}$0.74$ & \cellcolor{highlight}$78.01$ & \cellcolor{highlight}$86.65$ & \cellcolor{highlight}$80.69$ & \cellcolor{highlight}$68.09$ & \cellcolor{highlight}$-1.25$ \\
    & \cellcolor{highlight}\metricxqe & \cellcolor{highlight}$-1.79$ & \cellcolor{highlight}$80.27$ & \cellcolor{highlight}$89.66$ & \cellcolor{highlight}$83.17$ & \cellcolor{highlight}$-1.28$ & \cellcolor{highlight}$79.42$ & \cellcolor{highlight}$85.60$ & \cellcolor{highlight}$81.38$ & \cellcolor{highlight}$63.17$ & \cellcolor{highlight}$-1.46$ \\
    & \metricxxlqe & $-3.46$ & $80.32$ & $86.03$ & $82.13$ & $-3.19$ & $78.27$ & $85.08$ & $80.41$ & $63.25$ & $-1.64$ \\
    & \cellcolor{highlight}\gemba & \cellcolor{highlight}$-7.00$ & \cellcolor{highlight}$79.70$ & \cellcolor{highlight}$89.24$ & \cellcolor{highlight}$82.64$ & \cellcolor{highlight}$-5.00$ & \cellcolor{highlight}$79.01$ & \cellcolor{highlight}$83.83$ & \cellcolor{highlight}$80.55$ & \cellcolor{highlight}$65.22$ & \cellcolor{highlight}$-1.26$ \\
    & \mateseqe & $-6.00$ & $74.35$ & $95.01$ & $80.16$ & $-3.00$ & $75.99$ & $81.43$ & $77.72$ & $53.95$ & $-2.28$ \\
    & \cometqe & $-0.03$ & $75.62$ & $91.71$ & $80.32$ & $-0.00$ & $74.21$ & $86.43$ & $77.88$ & $61.06$ & $-1.70$ \\
    & \cometmqmqe & $0.08$ & $76.28$ & $89.22$ & $80.16$ & $0.09$ & $74.16$ & $87.36$ & $78.09$ & $52.57$ & $-2.32$ \\
    & \cometkiwi & $0.77$ & $80.14$ & $86.48$ & $82.14$ & $0.80$ & $80.02$ & $79.85$ & $79.96$ & $60.42$ & $-1.55$ \\
    & \cometkiwixl & $0.60$ & $77.83$ & $89.83$ & $81.46$ & $0.63$ & $76.90$ & $84.57$ & $79.30$ & $65.61$ & $-1.28$ \\
    \cmidrule(lr){2-12}
    \multirow{ 5 }{*}{\rotatebox{90}{\shortstack{\textsc{lexical} \\ \textsc{based}}}}
    & \bleu & $0.00$ & $71.31$ & $100.00$ & $78.85$ & $0.00$ & $67.89$ & $100.00$ & $76.03$ & $64.46$ & $-1.38$ \\
    & \chrf & $0.00$ & $71.31$ & $100.00$ & $78.85$ & $0.00$ & $67.89$ & $100.00$ & $76.03$ & $65.61$ & $-1.27$ \\
    & \ebleu & $0.02$ & $71.36$ & $99.94$ & $78.88$ & $0.02$ & $67.93$ & $99.94$ & $76.05$ & $65.29$ & $-1.32$ \\
    & \fbleu & $0.00$ & $71.31$ & $100.00$ & $78.85$ & $7.35$ & $68.71$ & $96.65$ & $76.04$ & $66.04$ & $-1.23$ \\
    & \tokengramf & $0.00$ & $71.31$ & $100.00$ & $78.85$ & $0.00$ & $67.89$ & $100.00$ & $76.03$ & $65.30$ & $-1.24$ \\
    \cmidrule(lr){2-12}
    \multirow{ 3 }{*}{\rotatebox{90}{\shortstack{\textsc{sentinel} \\ \textsc{metrics}}}}
    & \srconly & $-0.10$ & $74.72$ & $91.11$ & $79.48$ & $-0.01$ & $72.89$ & $88.48$ & $77.44$ & $53.09$ & $-2.35$ \\
    & \refonly & $-0.78$ & $73.28$ & $96.26$ & $79.62$ & $-0.78$ & $70.04$ & $96.66$ & $77.12$ & $53.09$ & $-2.35$ \\
    & \candonly & $-0.52$ & $74.56$ & $93.53$ & $79.97$ & $-0.50$ & $71.42$ & $93.81$ & $77.59$ & $45.30$ & $-3.04$ \\
    \cmidrule(lr){2-12}
    & \randombaseline & $-5.00$ & $71.32$ & $99.99$ & $78.85$ & $-5.00$ & $67.89$ & $99.99$ & $76.03$ & $53.51$ & $-2.08$ \\
    \bottomrule
    \end{NiceTabular}
}

\caption{Metrics' Precision, Recall, and \fscore in binary classification, distinguishing \good from \bad, and \perfect from \other translations. $\tau$ is selected to maximize the \fscore \textbf{on the test set}. In the last two columns, we report metrics' Precision in translation re-ranking and the average MQM score of the selected translations. The test set is \wmtth and the translation direction is \langpair{he}{en}. The metrics highlighted in grey are not openly available.}
\label{tab:performance-results-heen-apx}
\end{table*}

\begin{table*}[t]
\centering

    \begin{NiceTabular}{ll|rrrr|rrrr}[cell-space-limits=3pt]
    \toprule
     &  & \multicolumn{4}{c|}{\textbf{\good vs \bad}} & \multicolumn{4}{c}{\textbf{\perfect vs \other}} \\
    & \textbf{Metric} & \multicolumn{1}{c}{$\boldsymbol{\tau}$} & \multicolumn{1}{c}{\textbf{P}} & \multicolumn{1}{c}{\textbf{R}} & \multicolumn{1}{c|}{$\boldsymbol{F}$} & \multicolumn{1}{c}{$\boldsymbol{\tau}$} & \multicolumn{1}{c}{\textbf{P}} & \multicolumn{1}{c}{\textbf{R}} & \multicolumn{1}{c}{$\boldsymbol{F}$} \\
    \cmidrule(lr){2-10}
    \multirow{ 4 }{*}{\rotatebox{90}{\small \shortstack{\textsc{reference} \\ \textsc{based}}}}
    & \metricxxl & $-3.93$ & $76.58$ & $87.01$ & $79.77$ & $-2.97$ & $57.70$ & $88.80$ & $65.32$ \\
    & \comet & $0.77$ & $75.95$ & $75.28$ & $75.72$ & $0.79$ & $55.51$ & $74.57$ & $60.68$ \\
    & \bleurt & $0.61$ & $73.81$ & $79.92$ & $75.74$ & $0.64$ & $52.45$ & $76.89$ & $58.67$ \\
    & \bertscore & $0.92$ & $71.44$ & $63.39$ & $68.54$ & $0.93$ & $49.99$ & $59.81$ & $52.88$ \\
    \cmidrule(lr){2-10}
    \multirow{ 5 }{*}{\rotatebox{90}{\small \shortstack{\textsc{reference} \\ \textsc{free}}}}
    & \metricxxlqe & $-5.45$ & $73.36$ & $91.88$ & $78.64$ & $-3.54$ & $55.63$ & $90.32$ & $63.80$ \\
    & \cometmqmqe & $0.07$ & $72.57$ & $92.41$ & $78.17$ & $0.08$ & $52.59$ & $93.19$ & $61.53$ \\
    & \cometqe & $-0.02$ & $73.77$ & $85.81$ & $77.39$ & $-0.02$ & $50.54$ & $88.44$ & $58.96$ \\
    & \cometkiwi & $0.74$ & $75.57$ & $85.35$ & $78.58$ & $0.76$ & $53.38$ & $86.73$ & $61.23$ \\
    & \cometkiwixl & $0.62$ & $75.47$ & $83.37$ & $77.93$ & $0.64$ & $53.70$ & $85.03$ & $61.22$ \\
    \cmidrule(lr){2-10}
    \multirow{ 3 }{*}{\rotatebox{90}{\small \shortstack{\textsc{lexical} \\ \textsc{based}}}}
    & \fbleu & $4.86$ & $64.52$ & $89.01$ & $71.03$ & $4.86$ & $42.53$ & $89.35$ & $51.53$ \\
    & \bleu & $5.44$ & $64.78$ & $85.99$ & $70.58$ & $5.43$ & $42.73$ & $86.42$ & $51.39$ \\
    & \chrf & $2.08$ & $64.04$ & $99.84$ & $72.73$ & $2.08$ & $42.09$ & $99.77$ & $52.14$ \\
    \bottomrule
    \end{NiceTabular}

\caption{Metrics' Precision, Recall, and \fscore in binary classification, distinguish \good from \bad, and \perfect from \other translations. $\tau$ is selected to maximize the \fscore \textbf{on the development set}, i.e., \wmttw. The test set is \wmtth and the translation direction is \langpair{zh}{en}.}
\label{tab:performance-results-zhen-dev-apx}
\end{table*}

\begin{table*}[t]
\centering

       \begin{NiceTabular}{ll|rrrr|rrrr}[cell-space-limits=3pt]
    \toprule
     &  & \multicolumn{4}{c|}{\textbf{\good vs \bad}} & \multicolumn{4}{c}{\textbf{\perfect vs \other}} \\
    & \textbf{Metric} & \multicolumn{1}{c}{$\boldsymbol{\tau}$} & \multicolumn{1}{c}{\textbf{P}} & \multicolumn{1}{c}{\textbf{R}} & \multicolumn{1}{c|}{$\boldsymbol{F}$} & \multicolumn{1}{c}{$\boldsymbol{\tau}$} & \multicolumn{1}{c}{\textbf{P}} & \multicolumn{1}{c}{\textbf{R}} & \multicolumn{1}{c}{$\boldsymbol{F}$} \\
    \cmidrule(lr){2-10}
    \multirow{ 4 }{*}{\rotatebox{90}{\small \shortstack{\textsc{reference} \\ \textsc{based}}}}
    & \metricxxl & $-2.10$ & $73.10$ & $81.34$ & $75.65$ & $-1.10$ & $71.84$ & $71.41$ & $71.70$ \\
    & \comet & $0.72$ & $57.94$ & $94.05$ & $66.44$ & $0.80$ & $53.88$ & $84.32$ & $61.25$ \\
    & \bleurt & $0.60$ & $63.64$ & $87.97$ & $70.11$ & $0.66$ & $56.99$ & $82.09$ & $63.45$ \\
    & \bertscore & $0.68$ & $52.15$ & $99.84$ & $62.03$ & $0.68$ & $40.00$ & $99.79$ & $49.99$ \\
    \cmidrule(lr){2-10}
    \multirow{ 5 }{*}{\rotatebox{90}{\small \shortstack{\textsc{reference} \\ \textsc{free}}}}
    & \metricxxlqe & $-2.71$ & $70.68$ & $81.67$ & $74.00$ & $-1.66$ & $67.12$ & $76.24$ & $69.91$ \\
    & \cometmqmqe & $0.07$ & $72.32$ & $76.27$ & $73.59$ & $0.09$ & $69.07$ & $71.16$ & $69.75$ \\
    & \cometqe & $-0.05$ & $56.09$ & $94.26$ & $64.84$ & $-0.01$ & $47.75$ & $87.66$ & $56.29$ \\
    & \cometkiwi & $0.73$ & $60.98$ & $90.91$ & $68.50$ & $0.79$ & $55.90$ & $83.61$ & $62.84$ \\
    & \cometkiwixl & $0.52$ & $59.13$ & $91.29$ & $67.00$ & $0.62$ & $51.57$ & $86.45$ & $59.59$ \\
    \cmidrule(lr){2-10}
    \multirow{ 3 }{*}{\rotatebox{90}{\small \shortstack{\textsc{lexical} \\ \textsc{based}}}}
    & \fbleu & $3.18$ & $52.15$ & $99.74$ & $62.01$ & $3.67$ & $40.05$ & $99.79$ & $50.04$ \\
    & \bleu & $0.00$ & $52.05$ & $100.00$ & $61.95$ & $0.00$ & $39.95$ & $100.00$ & $49.94$ \\
    & \chrf & $0.00$ & $52.05$ & $100.00$ & $61.95$ & $0.00$ & $39.95$ & $100.00$ & $49.94$ \\
    \bottomrule
    \end{NiceTabular}

\caption{Metrics' Precision, Recall, and \fscore in binary classification, distinguish \good from \bad, and \perfect from \other translations. $\tau$ is selected to maximize the \fscore \textbf{on the development set}, i.e., \wmttw. The test set is \wmtth and the translation direction is \langpair{en}{de}.}
\label{tab:performance-results-ende-dev-apx}
\end{table*}

\section{Additional Figures} \label{apx:thresholds-stability}
\begin{figure*}[ht]
    \centering
    \resizebox{\linewidth}{!}{
    \includegraphics{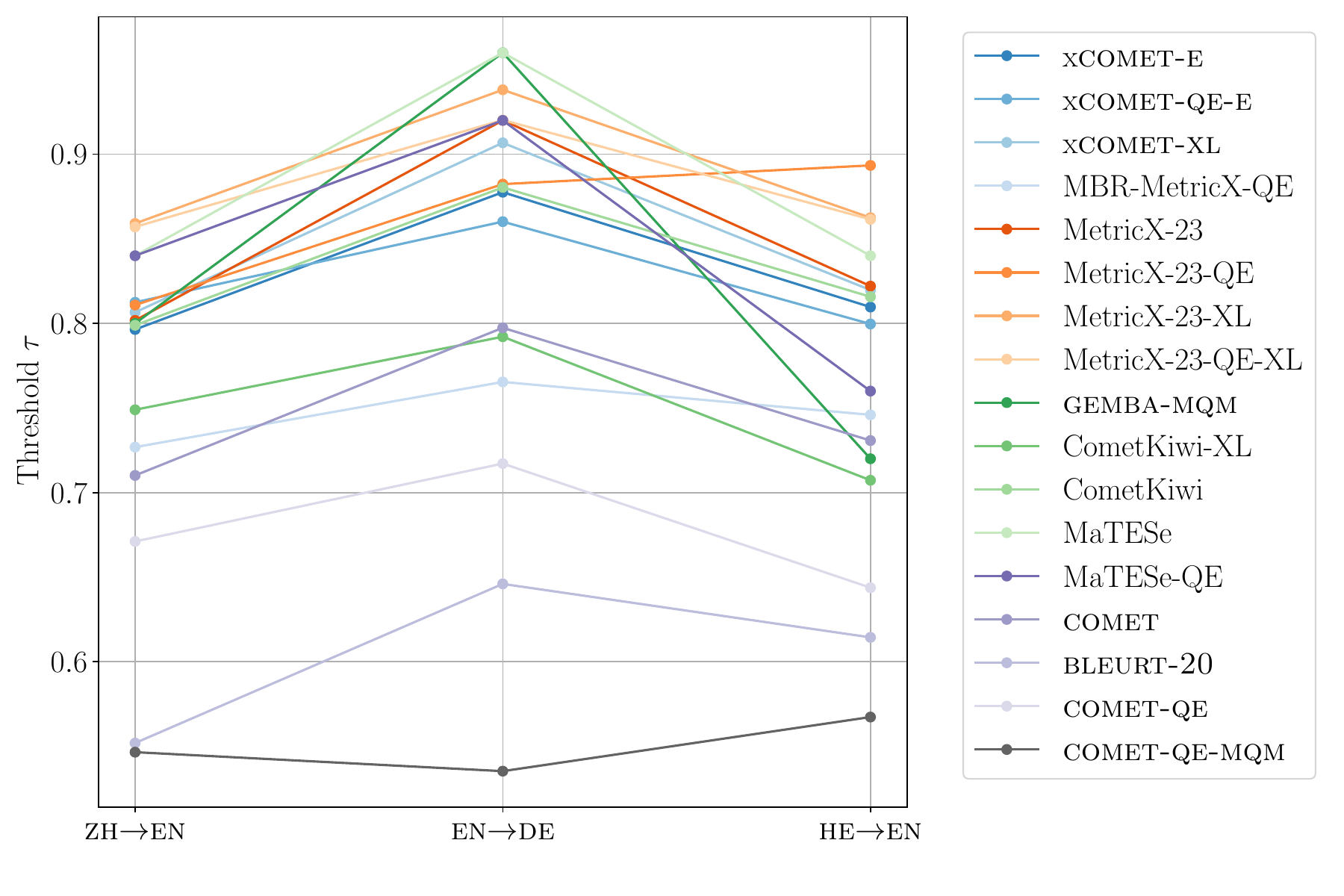}
    }
    \caption{Tested metrics' optimal threshold values across different language directions. The thresholds were selected to maximize the \fscore on the test set in the \good vs \bad binary classification scenario. Thresholds are normalized between $0$ and $1$ for improved clarity.}
    \label{fig:thresholds-full-stability-4.0}
\end{figure*}

\begin{figure*}[ht]
    \centering
    \resizebox{\linewidth}{!}{
    \includegraphics{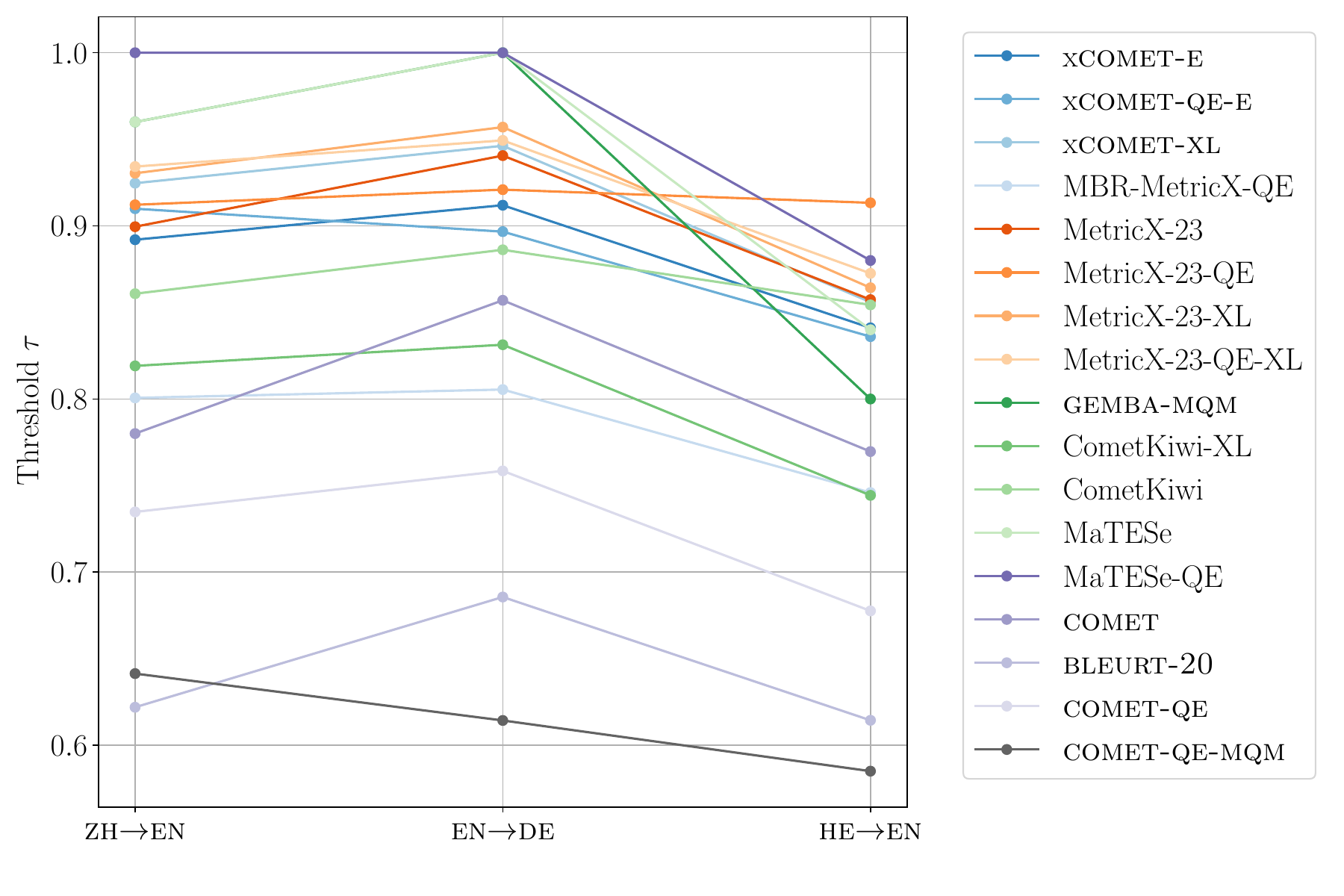}
    }
    \caption{Tested metrics' optimal threshold values across different language directions. The thresholds were selected to maximize the \fscore on the test set in the \perfect vs \other binary classification scenario. Thresholds are normalized between $0$ and $1$ for improved clarity.}
    \label{fig:thresholds-full-stability-1.0}
\end{figure*}

In Figures~\ref{fig:thresholds-full-stability-4.0} and \ref{fig:thresholds-full-stability-1.0}, we report metrics optimal threshold values across different language directions. The thresholds were selected to maximize the \fscore on the test set. 

\begin{figure*}
    \centering
    \includegraphics[width=1.2\columnwidth]{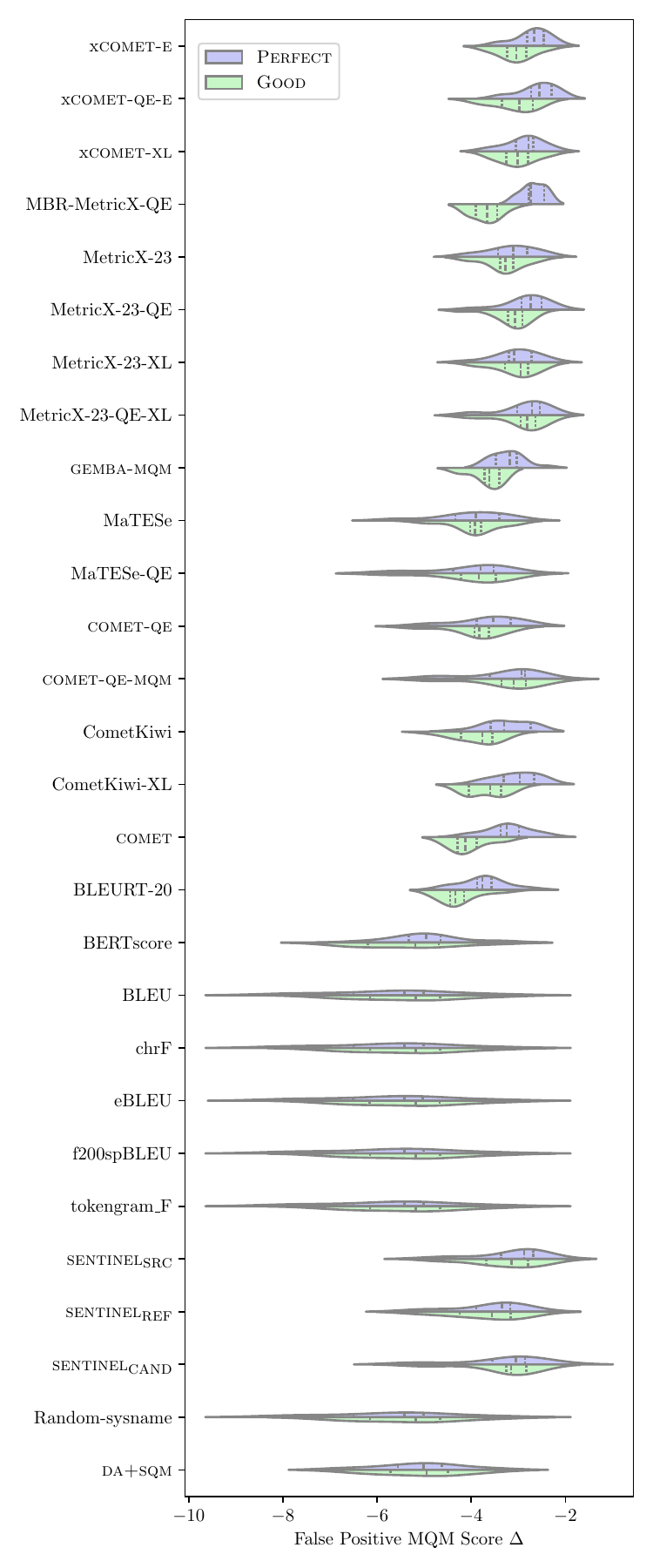}
    \caption{Distribution of the MQM score $\Delta$ between metrics' false positive MQM scores and human thresholds, i.e., $-4$ for \good and $-1$ for \perfect. The dataset is the \langpair{zh}{en} split of \wmtth.}
    \label{fig:fps-deltas-apx}
\end{figure*}

In Figure~\ref{fig:fps-deltas-apx} we report the $\Delta$ MQM score between the false positives and the human thresholds in the \good and \perfect translations classification scenario.

\section{\dasqm and MQM Correlation} \label{apx:dasqm-corr}
Tables~\ref{tab:dasqm-full-corr-zhen} and \ref{tab:dasqm-full-corr-ende} present the segment-level correlation between the tested metrics and MQM when considering \dasqm as a metric. We employ Pearson's $\rho$ and Kendall's $\tau$ correlation coefficients, and \acceq accuracy \cite{deutsch-etal-2023-ties}. As recommended by \citet{perrella-etal-2024-guardians}, we use \textit{Segment Grouping}, meaning that we compute these statistics on groups of translations of the same source text, and then average them. To enable a fair comparison between metrics and \dasqm, we restrict the evaluation datasets to the translations with available \dasqm annotations. 

\begin{table*}[t]
    \centering
    \begin{tabular}{lrrr}
        \toprule
        \textbf{Metric} & \multicolumn{1}{c}{$\boldsymbol{\tau}$} & \multicolumn{1}{c}{$\boldsymbol{\rho}$} & \multicolumn{1}{c}{\textbf{\acceq}} \\
        \midrule
        \gemba & $0.36$ & $0.43$ & $0.52$ \\
        \xcometens & $0.30$ & $0.42$ & $0.54$ \\
        \xcometensqe & $0.26$ & $0.37$ & $0.53$ \\
        \xcometxl & $0.26$ & $0.38$ & $0.52$ \\
        \mbrmetricxqe & $0.29$ & $0.43$ & $0.53$ \\
        \metricx & $0.26$ & $0.37$ & $0.53$ \\
        \metricxqe & $0.24$ & $0.35$ & $0.52$ \\
        \metricxxl & $0.25$ & $0.36$ & $0.52$ \\
        \cometkiwi & $0.25$ & $0.37$ & $0.52$ \\
        \cometkiwixl & $0.25$ & $0.38$ & $0.52$ \\
        \comet & $0.25$ & $0.36$ & $0.51$ \\
        \bleurt & $0.26$ & $0.37$ & $0.52$ \\
        \matese & $0.27$ & $0.33$ & $0.48$ \\
        \cometmqmqe & $0.16$ & $0.21$ & $0.48$ \\
        \mateseqe & $0.21$ & $0.24$ & $0.44$ \\
        \dasqm & $0.11$ & $0.20$ & $0.42$ \\
        \randombaseline & $0.02$ & $0.02$ & $0.38$ \\
        \bottomrule
    \end{tabular}
    \caption{Kendall $\tau$ and Pearson $\rho$ correlation coefficients, and \acceq accuracy \cite{deutsch-etal-2023-ties}, measured between the \dasqm- and MQM-based annotations, and between MT metrics and MQM. The data is the intersection between \wmtth and \wmtthda. The language direction is \langpair{zh}{en}.}
    \label{tab:dasqm-full-corr-zhen}
\end{table*}
\begin{table*}[]
    \centering
    \begin{tabular}{lrrr}
        \toprule
        \textbf{Metric} & \multicolumn{1}{c}{$\boldsymbol{\tau}$} & \multicolumn{1}{c}{$\boldsymbol{\rho}$} & \multicolumn{1}{c}{\textbf{\acceq}} \\
        \midrule
        \gemba & $0.40$ & $0.48$ & $0.57$ \\
        \mbrmetricxqe & $0.40$ & $0.54$ & $0.58$ \\
        \xcometens & $0.38$ & $0.54$ & $0.60$ \\
        \xcometxl & $0.37$ & $0.51$ & $0.60$ \\
        \metricx & $0.37$ & $0.51$ & $0.60$ \\
        \comet & $0.37$ & $0.51$ & $0.58$ \\
        \bleurt & $0.37$ & $0.49$ & $0.57$ \\
        \metricxxl & $0.36$ & $0.49$ & $0.59$ \\
        \xcometensqe & $0.36$ & $0.51$ & $0.59$ \\
        \metricxqe & $0.36$ & $0.51$ & $0.60$ \\
        \cometqe & $0.35$ & $0.47$ & $0.57$ \\
        \metricxxlqe & $0.35$ & $0.45$ & $0.59$ \\
        \cometkiwixl & $0.35$ & $0.50$ & $0.57$ \\
        \cometkiwi & $0.33$ & $0.46$ & $0.57$ \\
        \cometmqmqe & $0.29$ & $0.39$ & $0.54$ \\
        \matese & $0.29$ & $0.33$ & $0.53$ \\
        \mateseqe & $0.28$ & $0.34$ & $0.52$ \\
        \dasqm & $0.17$ & $0.29$ & $0.46$ \\
        \randombaseline & $0.08$ & $0.12$ & $0.41$ \\
        \bottomrule
    \end{tabular}
    \caption{Kendall $\tau$ and Pearson $\rho$ correlation coefficients, and \acceq accuracy \cite{deutsch-etal-2023-ties}, measured between the \dasqm- and MQM-based annotations, and between MT metrics and MQM. The data is the intersection between \wmtth and \wmtthda. The language direction is \langpair{en}{de}.}
    \label{tab:dasqm-full-corr-ende}
\end{table*}

\end{document}